\documentclass[10pt,letterpaper]{article}
\usepackage[top=0.85in,left=2.75in,footskip=0.75in,marginparwidth=2in]{geometry}

\usepackage[utf8]{inputenc}

\usepackage{cite}

\usepackage{nameref,hyperref}

\usepackage{microtype}
\DisableLigatures[f]{encoding = *, family = * }

\raggedright
\usepackage{changepage}

\usepackage[aboveskip=1pt,labelfont=bf,labelsep=period,singlelinecheck=off]{caption}

\makeatletter
\renewcommand{\@biblabel}[1]{\quad#1.}
\makeatother

\usepackage{lastpage,fancyhdr,graphicx}
\usepackage{epstopdf}
\fancyheadoffset[L]{2.25in}
\fancyfootoffset[L]{2.25in}

\usepackage{color}

\definecolor{Gray}{gray}{.25}

\usepackage{graphicx}

\usepackage{sidecap}

\usepackage{wrapfig}
\usepackage[pscoord]{eso-pic}
\usepackage[fulladjust]{marginnote}
\reversemarginpar

\usepackage{booktabs}

\usepackage{subcaption}

\usepackage{float}

\begin{document}
\vspace*{0.35in}

\begin{flushleft}
{\Large
\textbf\newline{Regional biases in image geolocation estimation: a case study with the SenseCity Africa dataset}
}
\newline
\\
Ximena Salgado Uribe\textsuperscript{1,*},
Martí Bosch\textsuperscript{1},
Jérôme Chenal\textsuperscript{1,2},
\\
\bigskip
\bf{1} Urban and Regional Planning Community (CEAT), École Polytechnique Fédérale de Lausanne (EPFL), Lausanne, Switzerland
\\
\bf{2} Center of Urban Systems (CUS), Mohamed VI Polytechnic University (UM6P), Lot 660, Hay Moulay Rachid, Ben Guerir 43150, Morocco
\\
\bigskip
* ximena.salgadouribe@epfl.ch

\end{flushleft}

\section*{Abstract}
  Advances in Artificial Intelligence are challenged by the biases rooted in the datasets used to train the models.
  In image geolocation estimation, models are mostly trained using data from specific geographic regions, notably the Western world, and as a result, they may struggle to comprehend the complexities of underrepresented regions.
  To assess this issue, we apply a state-of-the art image geolocation estimation model (ISNs) to a crowd-sourced dataset of geolocated images from the African continent (SCA100), and then explore the regional and socioeconomic biases underlying the model's predictions.
  Our findings show that the ISNs model tends to over-predict image locations in high-income countries of the Western world, which is consistent with the geographic distribution of its training data, i.e., the IM2GPS3k dataset.
  Accordingly, when compared to the IM2GPS3k benchmark, the accuracy of the ISNs model notably decreases at all scales.
  Additionally, we cluster images of the SCA100 dataset based on how accurately they are predicted by the ISNs model and show the model's difficulties in correctly predicting the locations of images in low income regions, especially in Sub-Saharan Africa.
  Therefore, our results suggest that using IM2GPS3k as a training set and benchmark for image geolocation estimation and other computer vision models overlooks its potential application in the African context.


\section*{Introduction}
In African cities, the digitization process is happening faster and is more ubiquitous than one might expect. At the same time, the remarkable advancement of artificial intelligence in computer vision (CV) paves the way for technological and societal progress. However, the existing data used to train these models often reflects societal biases and systemic inequalities. By neglecting to capture the realities of African cities, biases in CV algorithms can perpetuate stereotypes and reinforce disparities. For instance, if the training data primarily consists of images from specific neighborhoods or regions, resulting models may struggle to interpret unique cultural nuances in other areas.
This can lead to misinterpretations in applications and services powered by these algorithms.

Image geolocation estimation is a CV problem which consists on determining the geographic location of a given image. This task has a wide-range of significant applications such as tourism, navigation systems, visual investigation in journalism, disaster management, automated image retrieval or the support and enhancement of historical image archives.
However, image geolocalization estimation poses many challenges, starting from characteristics of the images such as the variety of viewing angles, times of the day, weather and lignthing conditions; to the specific nature of the location itself, especially the lack of identifiable landmarks or the presence of multiple similar scenes around the world.
The above issues are exacerbated in regions where training data is scarce, such as rural areas and developing countries.
In light of the above, the objective of this article is to evaluate the accuracy of the a state-of-the-art image geolocation estimation model named Individual Scene Networks (ISNs) \cite{muller2018geolocation} on SenseCity Africa 100 (SCA100) a crowd-sourced dataset of images of African urban and rural scenes.
Additionally, we assess the geographic patterns of the model's mispredictions by exploring regional distinctions (which integrate demographic, cultural, political, infrastructure, and geographical aspects) and country income groups.

In essence, our research aims to contribute to the broader discourse on addressing biases in CV algorithms, particularly in the context of the Global South.
This paper's relevance resonates with academic researchers, policymakers, educators, and stakeholders committed to fostering inclusive, equitable, and representative AI technologies that resonate with diverse global populations.




The rest of the article is structured as follows: a review of the background on image geolocation estimation and biases in CV, followed by an examination of the dataset and the methods. Subsequently, we discuss our findings and their implications. Finally, we conclude by addressing limitations of the current and outlining directions for future work. 

\section*{Background}

\subsection*{Image gelocalization prediction}

As reviewed in the introduction, determining the location of an image remains a challenge despite the vast availability of billions of geo-tagged images.

Initial geolocation estimation works focused on restricted tasks like landmark recognition \cite{quack2008world,zheng2009tour,avrithis2010retrieving,chen2011city}, or focused on predictions within specific, relatively small geographical regions such as cities with street view imagery \cite{zamir2014image,kim2015predicting,tian2017cross}.
Similar studies focused on predicting geolocations in natural, non-urban settings, such as beaches \cite{cao2012bluefinder} or mountainous areas \cite{baatz2012large,saurer2016image}.

On the other hand, several works have attempted global-scale image geolocation prediction, encompassing unrestricted geographical areas that span the entire planet.  
The seminal works of Hays and Efros \cite{hays2008im2gps,hays2015large} introduced the IM2GPS dataset, featuring over 6 million geolocated images from the Internet. Subsequently, Vo et al. \cite{vo2017revisiting} revisited image geolocation estimation on IM2GPS by leveraging advances on Convolutional Neural Networks (CNNs). Similarly, the PlaNet model \cite{seo2018cplanet} also proposed a CNNs approach that reformulated image geolocation estimation as a classification task (replacing traditional nearest neighbor searches) by partitioning the world into a grid so that the key task is to assign query image to a grid cell. Following this idea, Müller et al. \cite{muller2018geolocation} introduced the (ISNs) model, which aims at reducing the number of candidate cells by incorporating contextual information about the environment at different scales.
In more recent developments, the Pigeon model \cite{haas2023pigeon} further expanded on earlier efforts by integrating climate data as an augmentation factor, acknowledging the impact of climate and environment context on various visual clues, leading to refined global-scale geolocation predictions.

\subsection*{Biases in training data}

Another key challenge of global image geolocation estimation is that the availability of training data is very unevenly distributed accross the world.
More broadly, the acknowledgment of undesirable biases in CV models has become a well-know fact. Torralba and Efros \cite{torralba2011unbiased} were among the first to thoroughly explore this issue, showing that large-scale datasets and the models trained on them tend to magnify biases because of the underrepresentation of minority groups and diverse cultures.

Since then, various works have highlighted gender, racial and georgraphic biases. For instance, the study of Buolamwini and Gebru \cite{buolamwini2018gender} showed that in facial recongition models black women had higher error rate than white men. Similarly, in object classification, certain objects, often linked to domestic or traditional roles, become associated particular with female gender \cite{zhao2017men}. Similarly, Hendricks et al. \cite{hendricks2018women} showcased gender imbalances in image captioning within the MSCOCO dataset \cite{lin2014microsoft}. Moreover, in the domain of text-to-image generation, studies demonstrated that images generated for everyday situations such as parks, food, and weddings are more closely aligned with the United States and Germany when using default location-neutral prompts \cite{naik2023social}, highlighting the influence of social and geographic factors on AI-generated content.

Similar issues have been identified in image geolocation prediction, where even though training datasets can comprise thousands of images, the absence of ordinary streets or even whole cities, plus the overrepresentation of attractive landmarks and tourist spots introduce a skewed perspective.
For instance, Hays and Efros \cite{hays2015large, hays2008im2gps} note that around 4.9\% of images in the IM2GPS database are located within 200km of London.
Similarly, although the "Mapillary Street-Level Sequences" dataset \cite{warburg2020mapillary} features more than 1.6 million images, a mere 2.9\% (5K) comes from Africa.
This reality raises an unavoidable concern, namely the lack of data for numerous locations across the globe, which can results in an inability to accurately localize queries within extensive regions of South America, Africa, and Central Asia \cite{shankar2017no}.

\section*{Materials and Methods}

This study focuses on the state-of-the-art ISNs model by Müller et al. \cite{muller2018geolocation} because it is the one with the best performance on image geolocation estimation over the IM2GPS dataset\footnote{By 21 January 2024, the ISNs model outperforms all other models in the ``Photo geolocation estimation on Im2GPS'' task according to Papers with Code. See https://paperswithcode.com/sota/photo-geolocation-estimation-on-im2gps (accessed 21 January 2023)}. In line with other models \cite{weyand2016planet,vo2017revisiting,muller2018geolocation}, we evaluate the ISNs model accuracy for the IM2GPS3k and SCA100 at five levels of granularity, namely street (1 km), city (25 km), region (200 km), country (750 km), and continent level (2500 km).

\begin{figure}[ht]
  \centering
  \begin{subfigure}[b]{\textwidth}
    \centering
    \includegraphics[width=.19\textwidth]{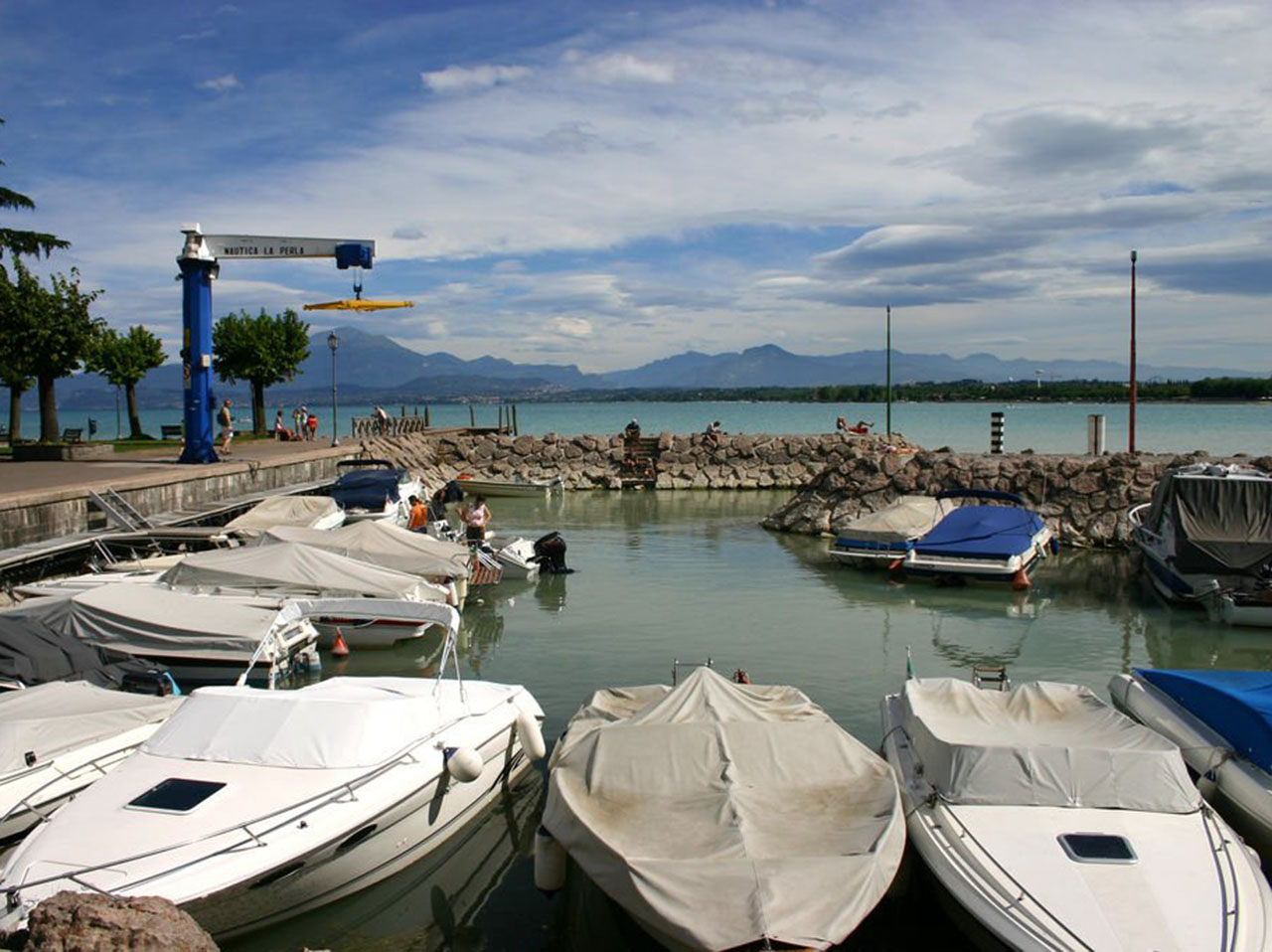}
    \includegraphics[width=.19\textwidth]{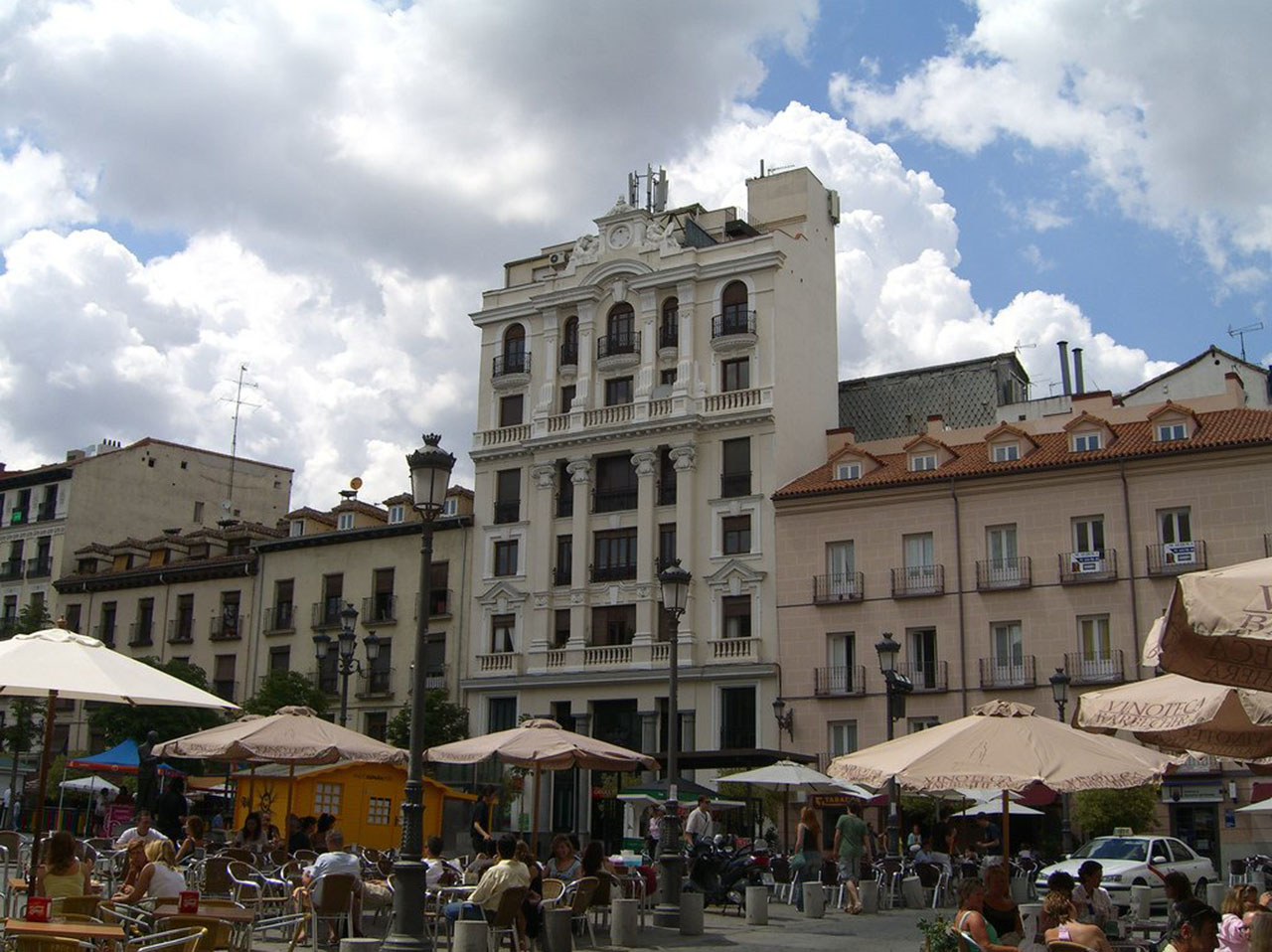}
    \includegraphics[width=.19\textwidth]{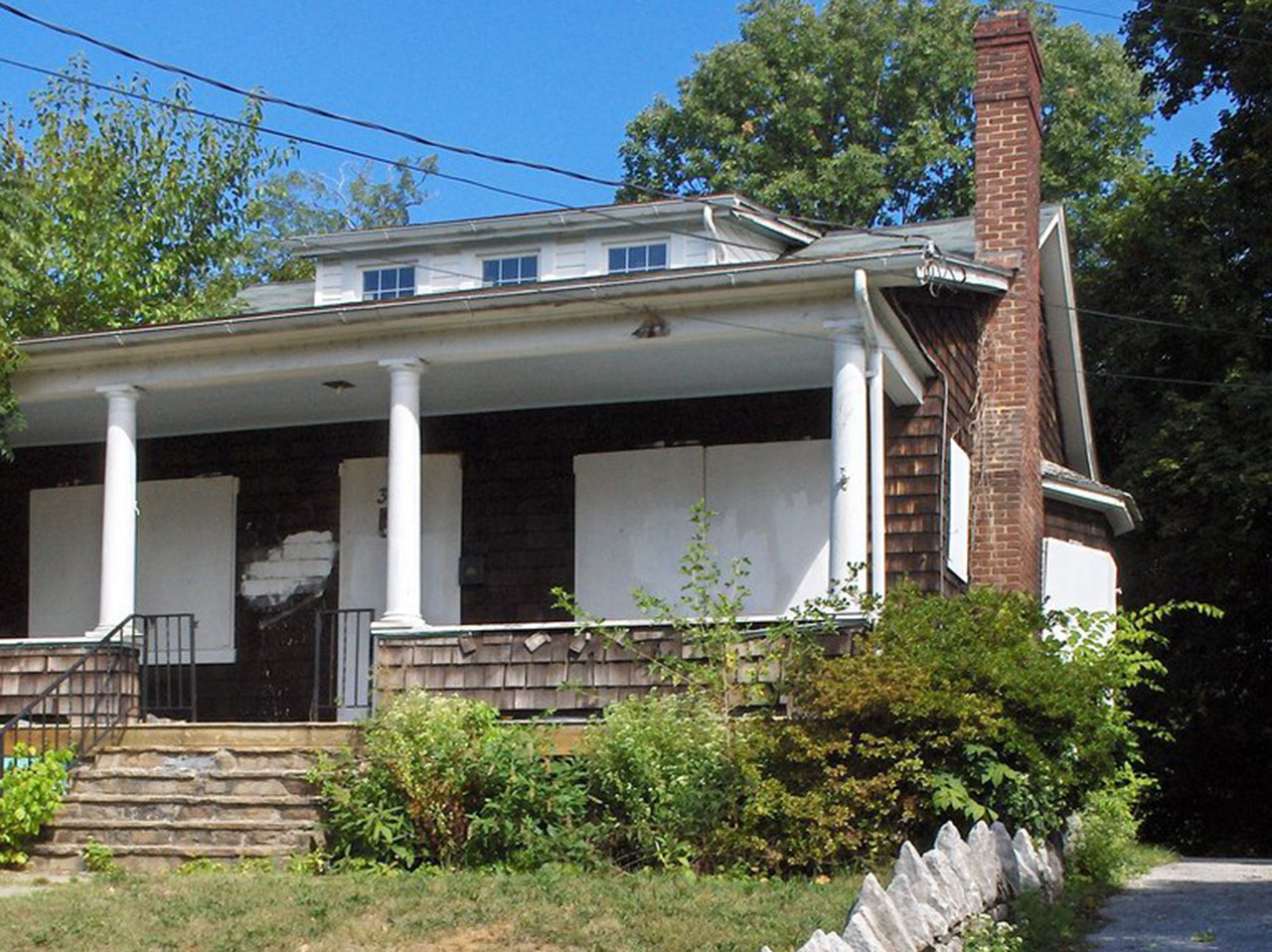}
    \includegraphics[width=.19\textwidth]{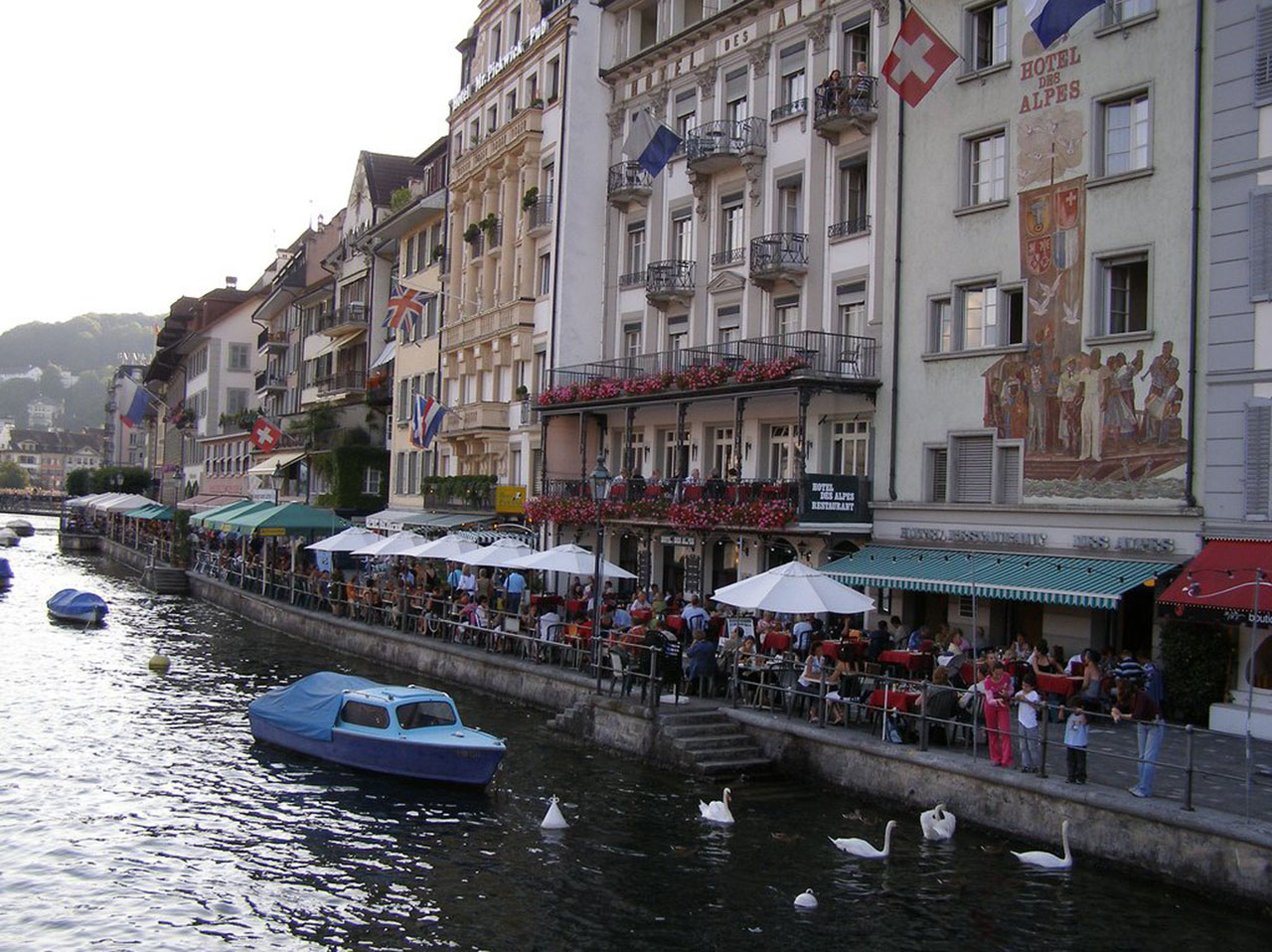}
    \includegraphics[width=.19\textwidth]{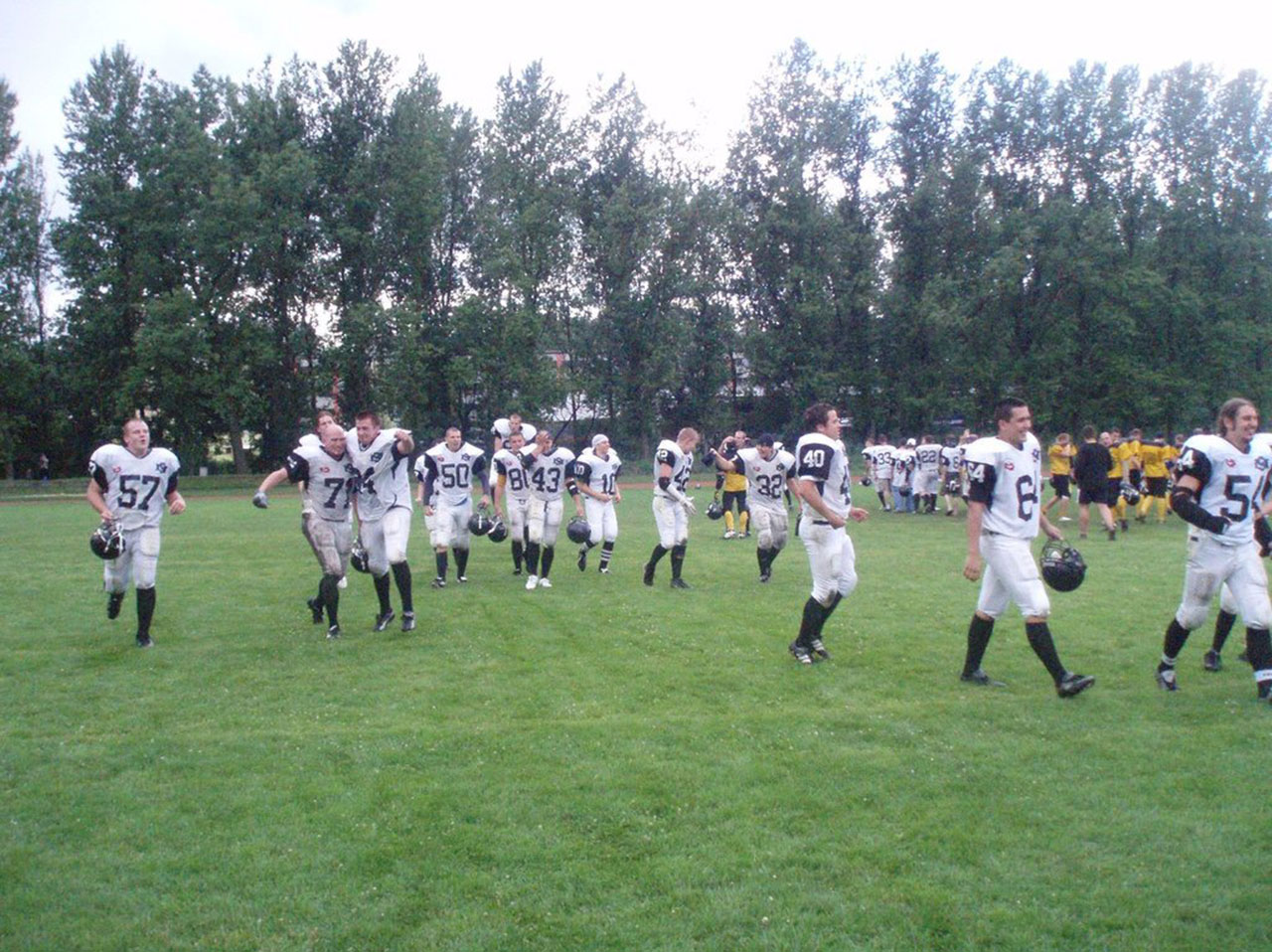}
    \includegraphics[width=.19\textwidth]{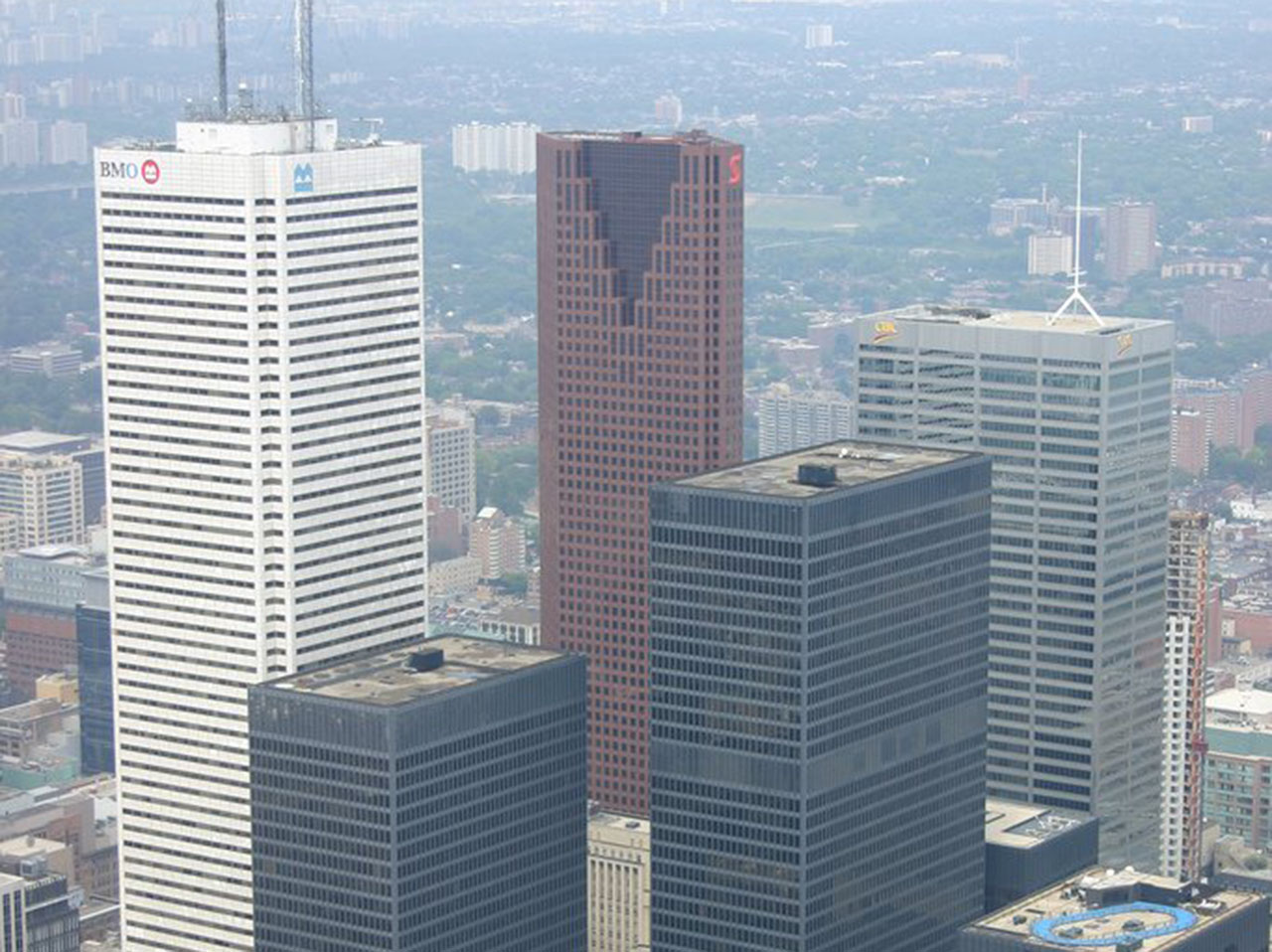}
    \includegraphics[width=.19\textwidth]{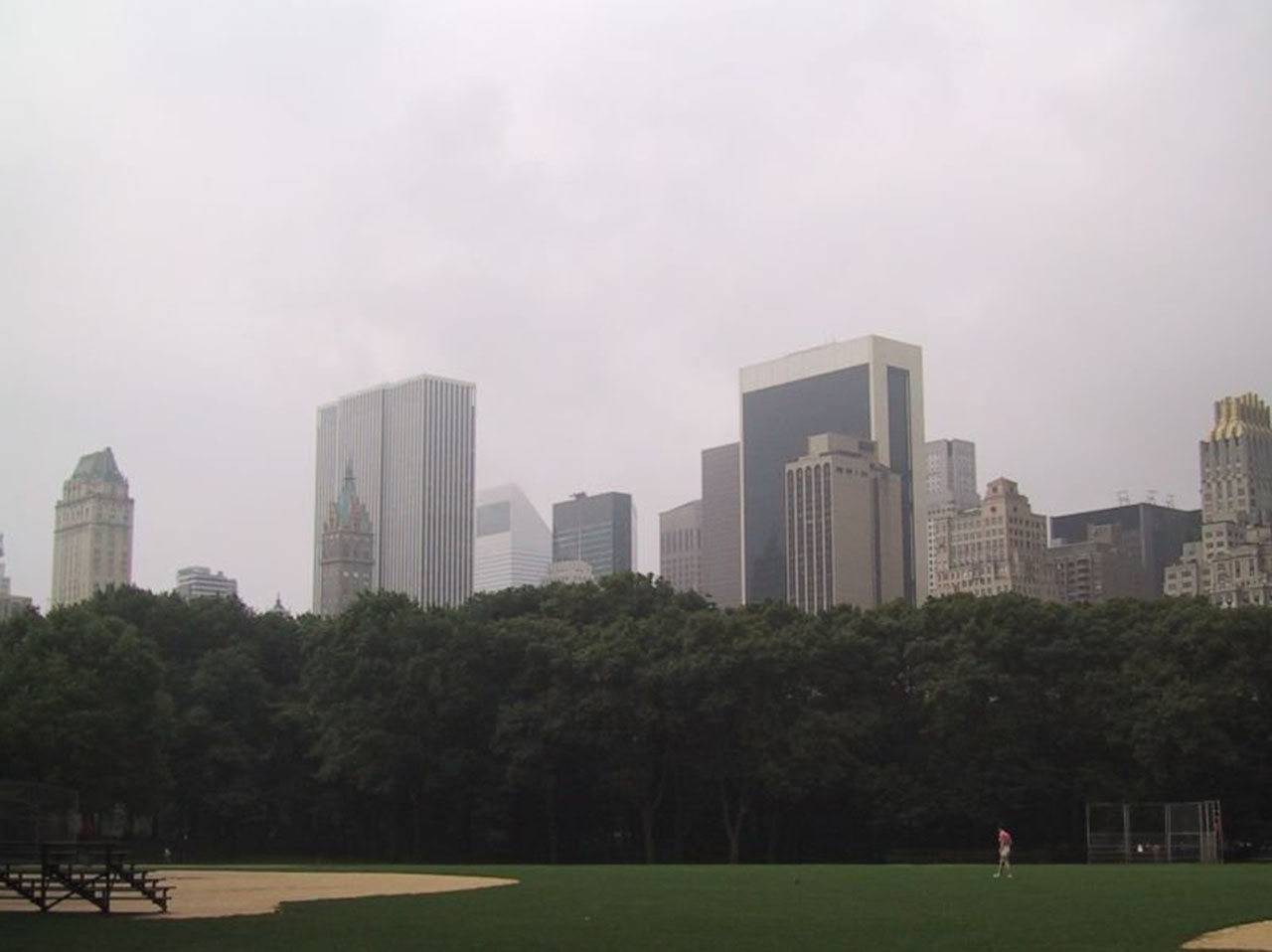}
    \includegraphics[width=.19\textwidth]{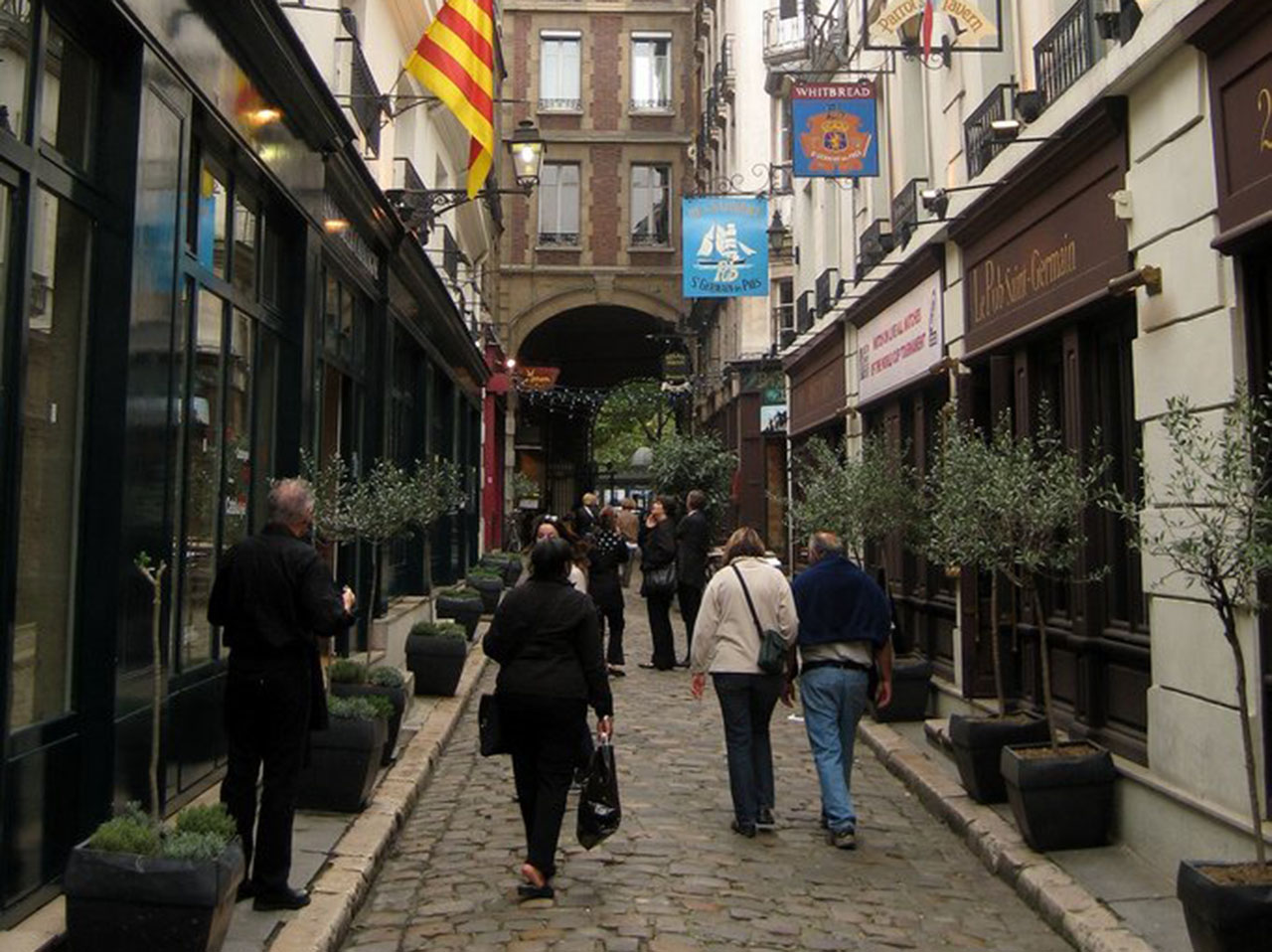}
    \includegraphics[width=.19\textwidth]{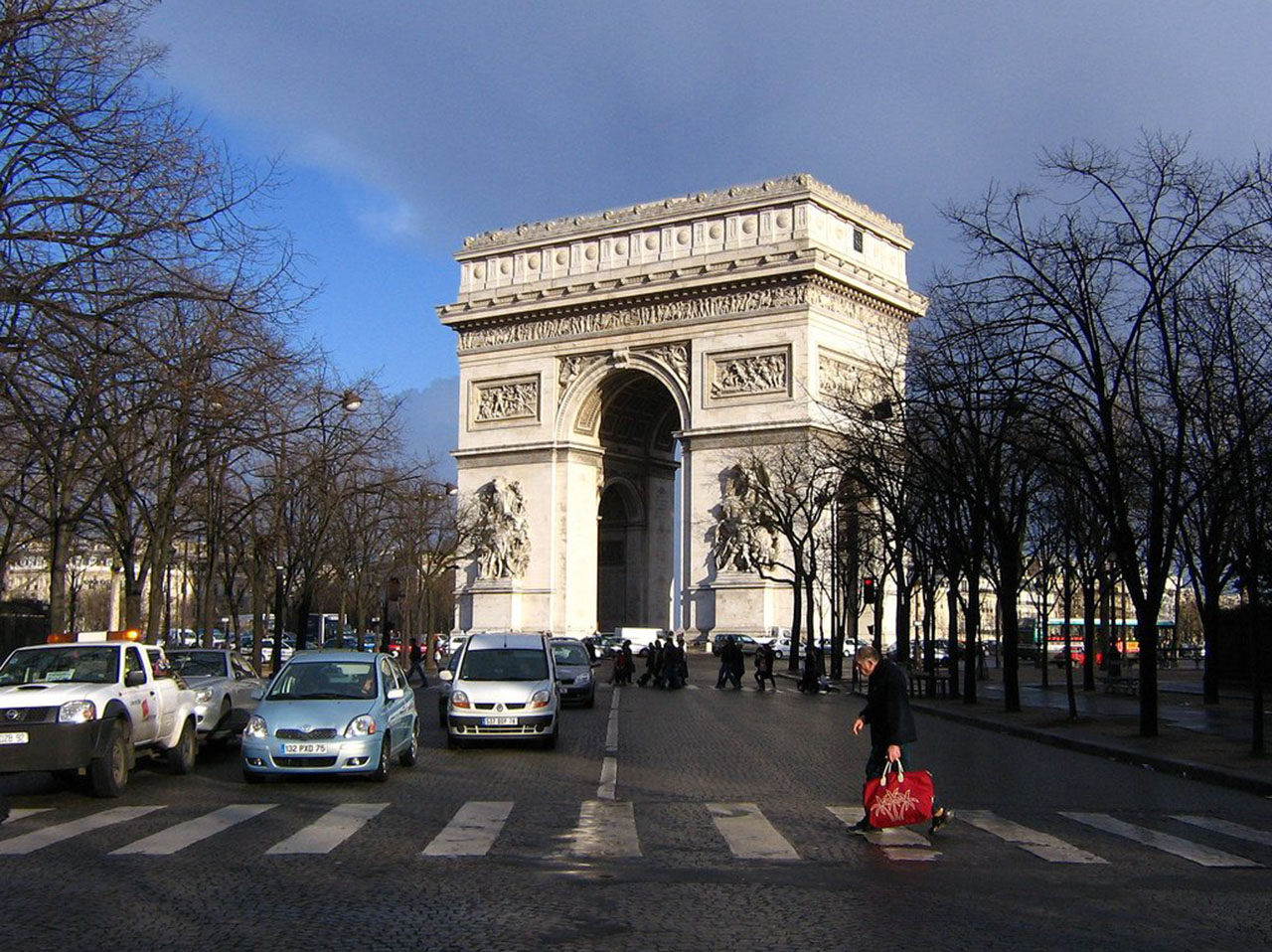}
    \includegraphics[width=.19\textwidth]{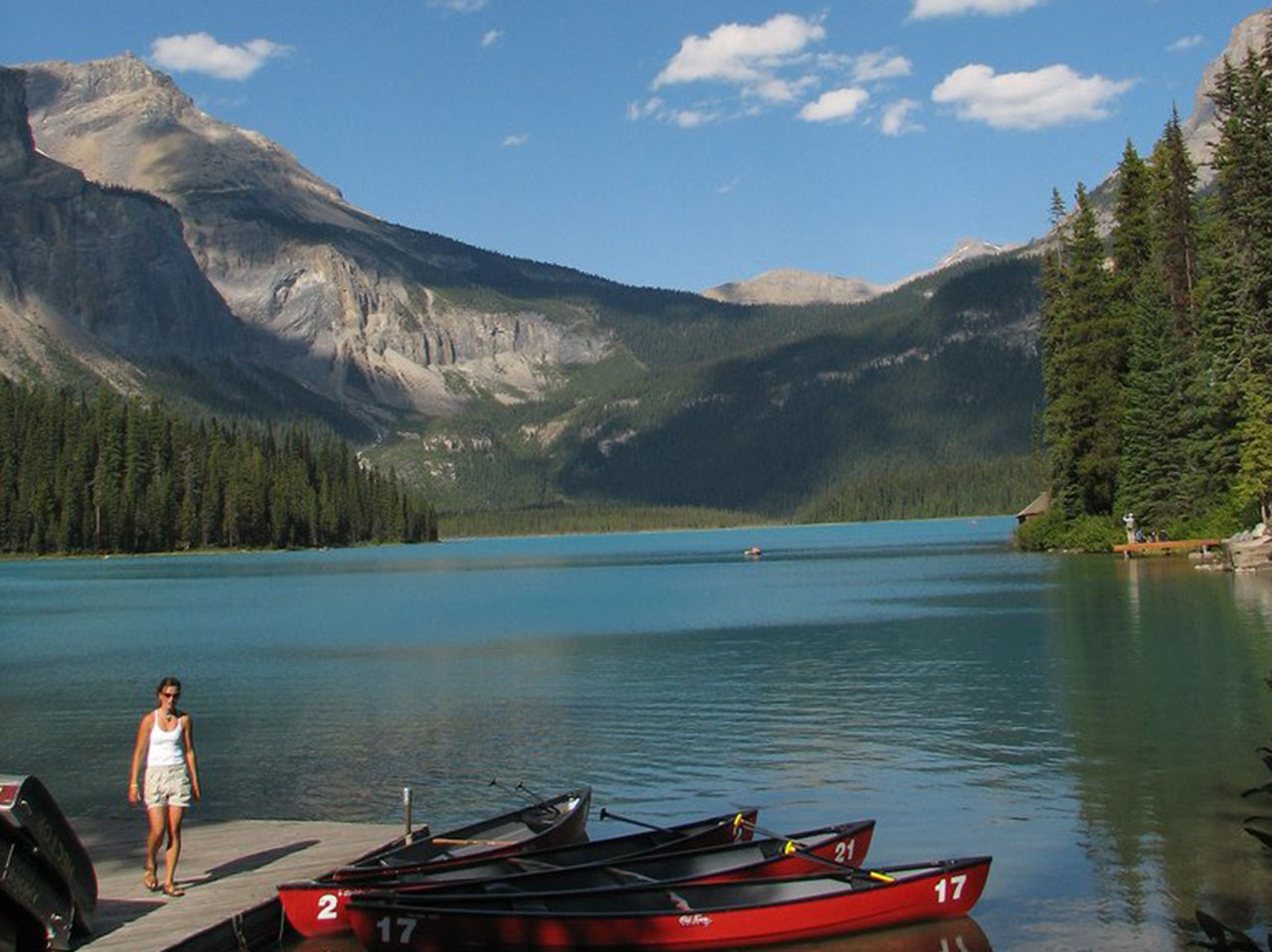}    
    \caption{\label{fig:im2gps} IM2GPS}
  \end{subfigure}
  \par\bigskip
  \begin{subfigure}[b]{\textwidth}
    \centering
    \includegraphics[width=.19\textwidth]{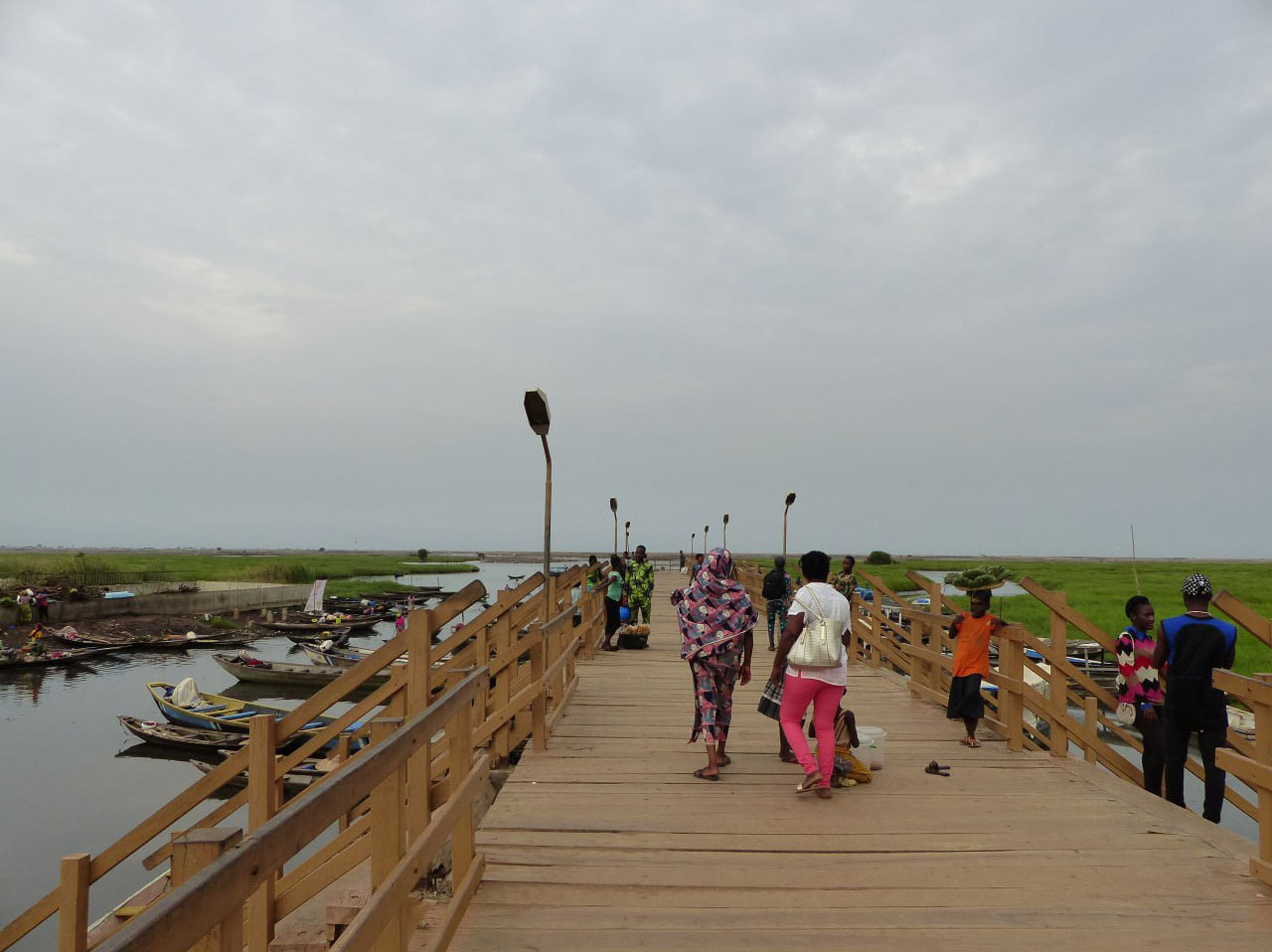}
    \includegraphics[width=.19\textwidth]{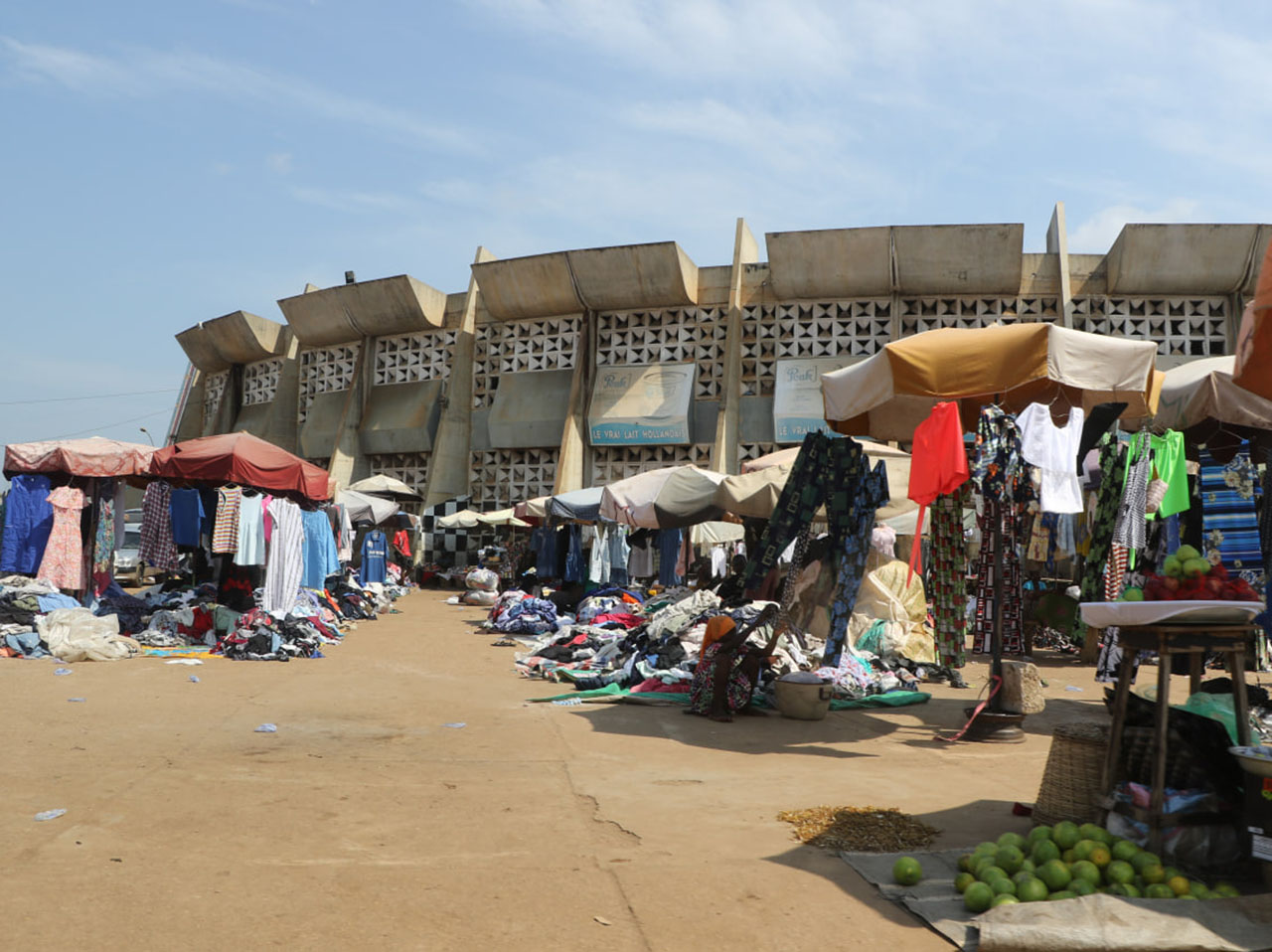}
    \includegraphics[width=.19\textwidth]{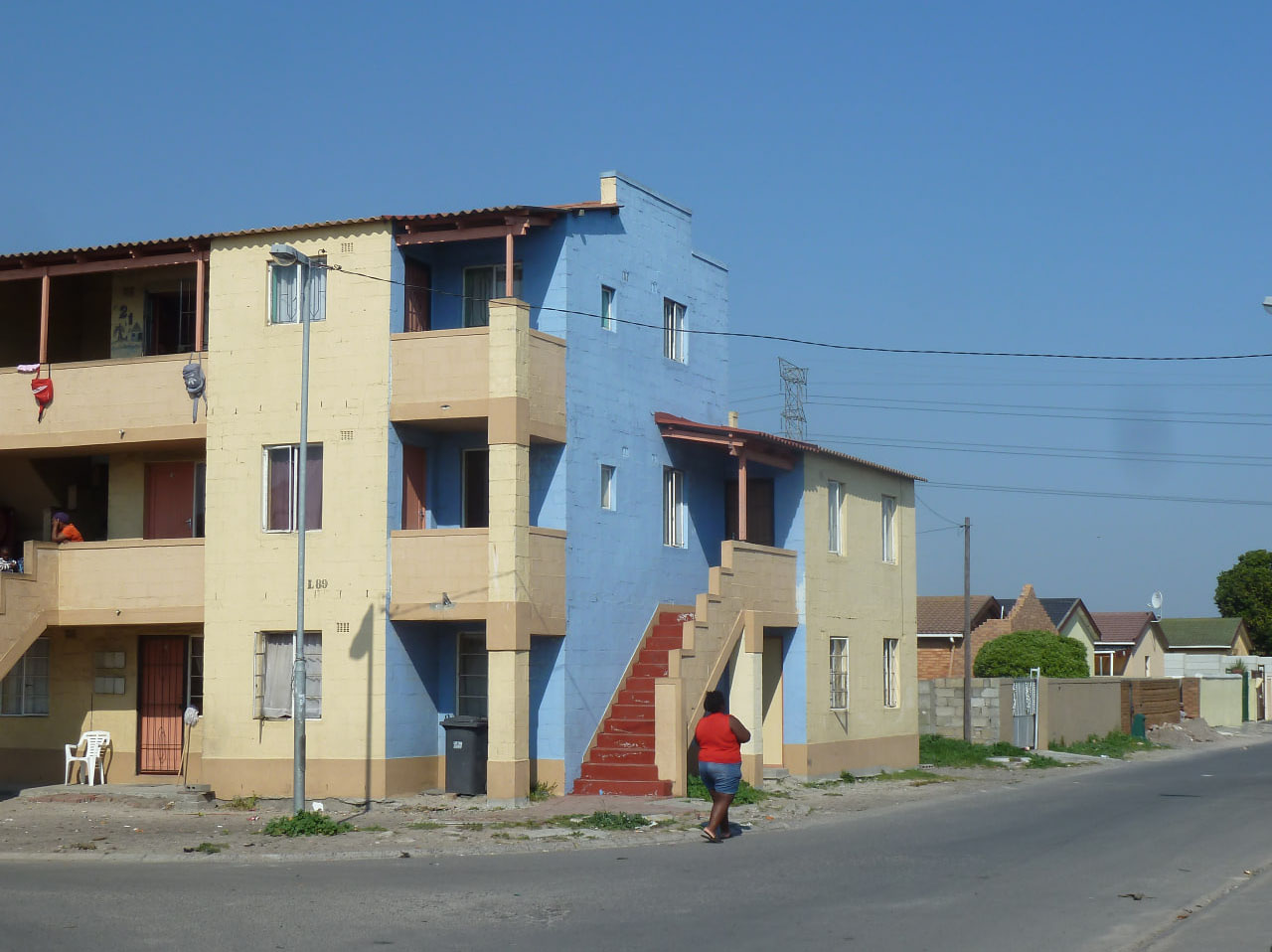}
    \includegraphics[width=.19\textwidth]{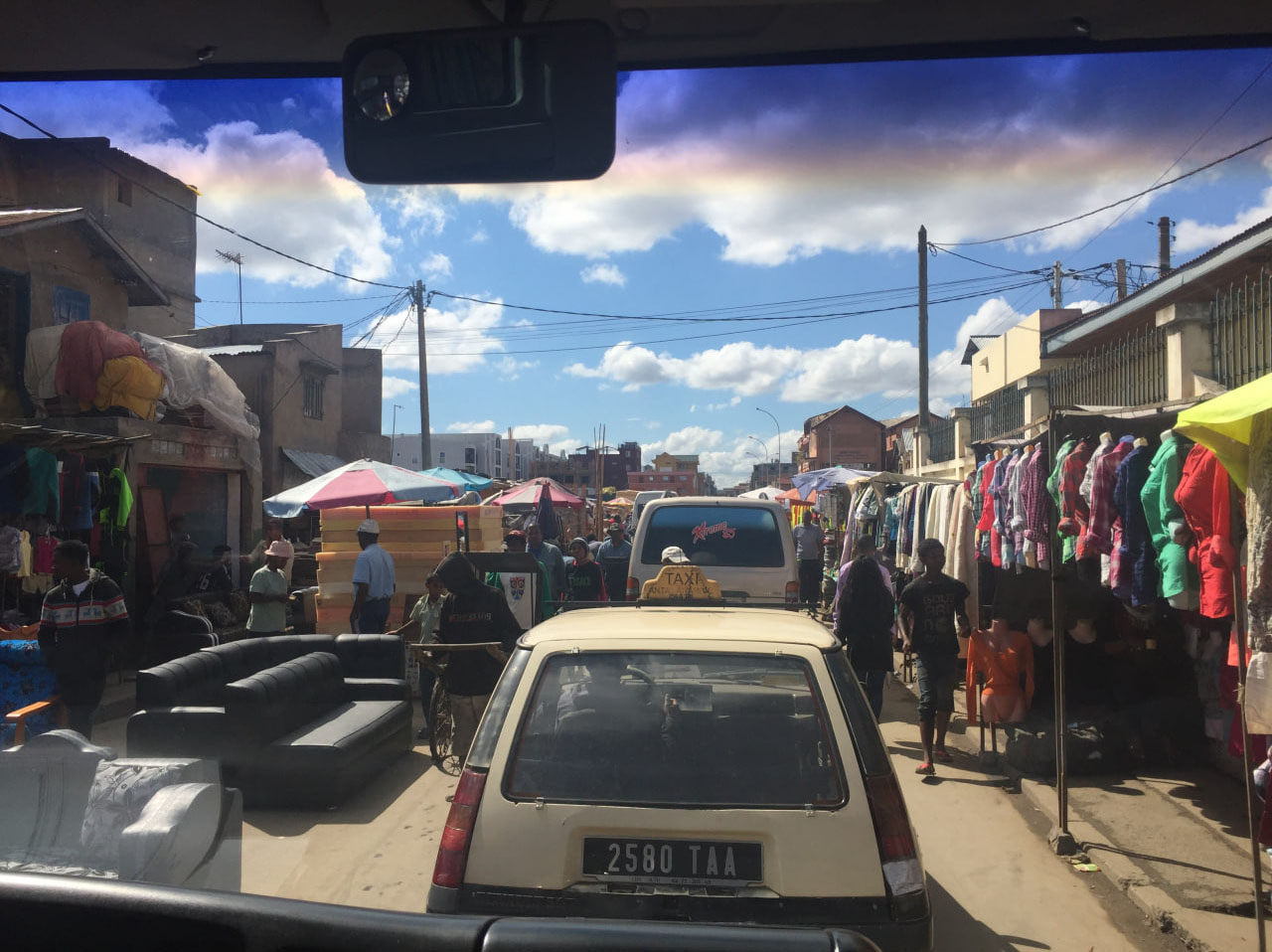}
    \includegraphics[width=.19\textwidth]{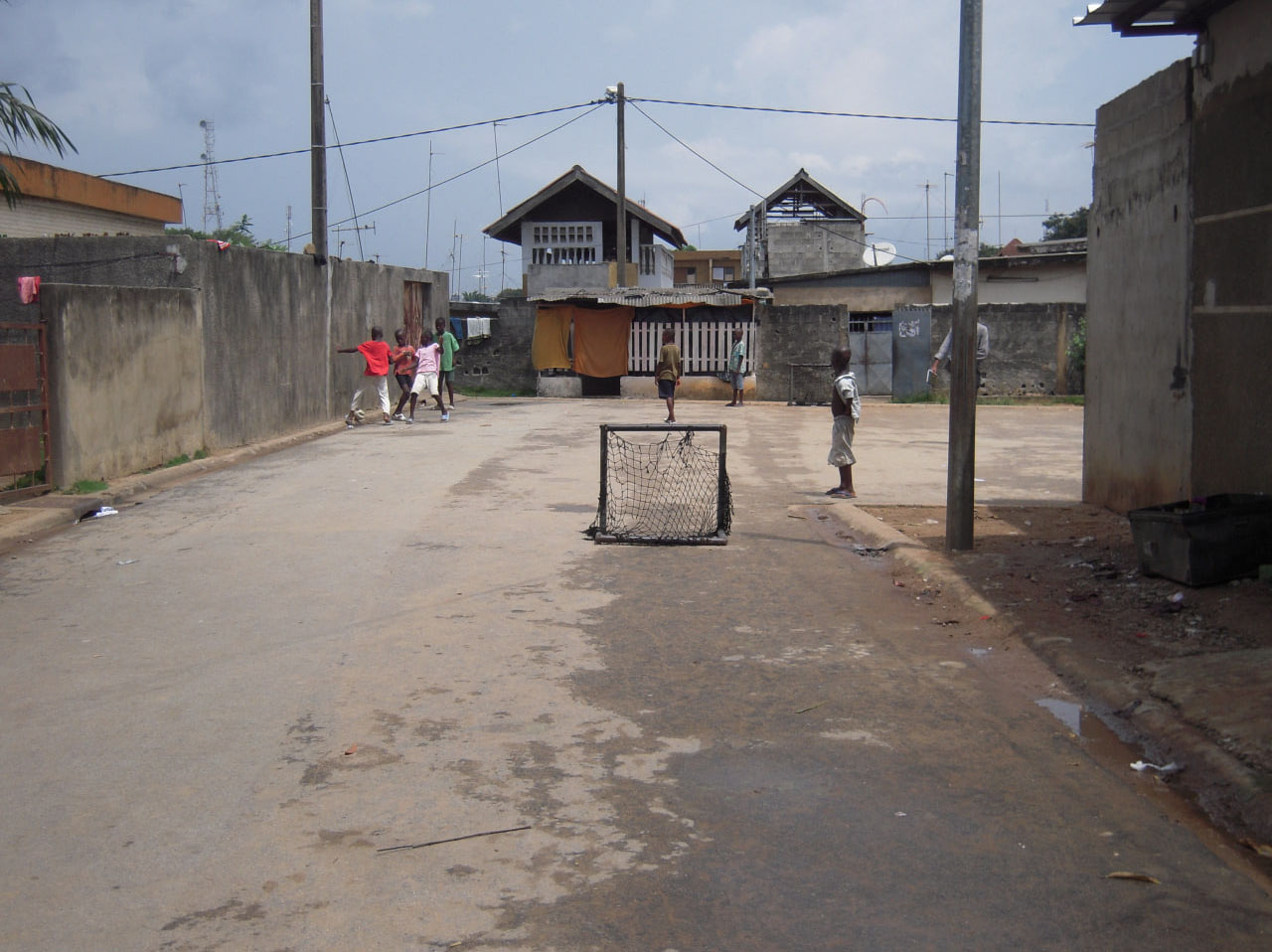}
    \includegraphics[width=.19\textwidth]{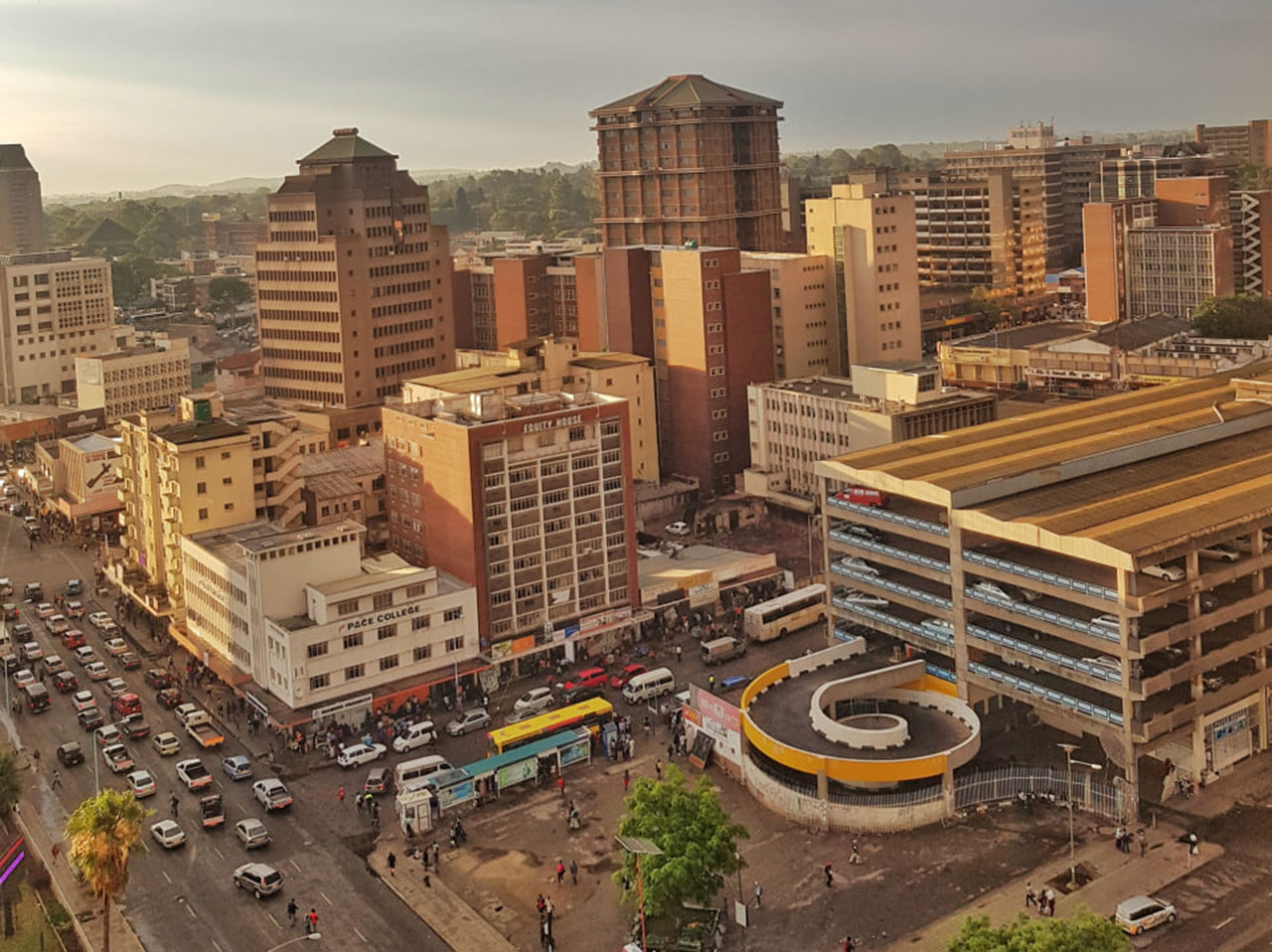}
    \includegraphics[width=.19\textwidth]{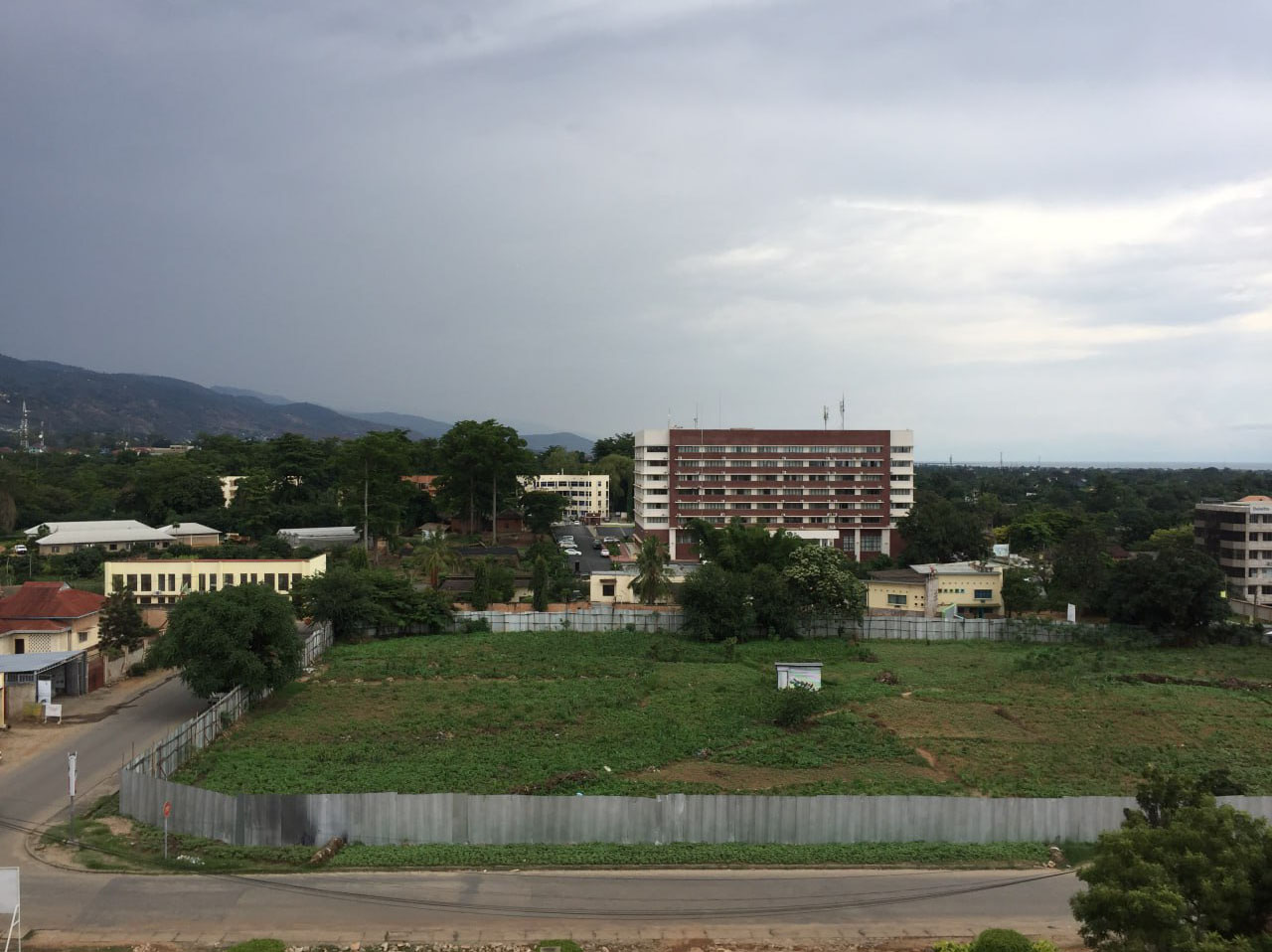}
    \includegraphics[width=.19\textwidth]{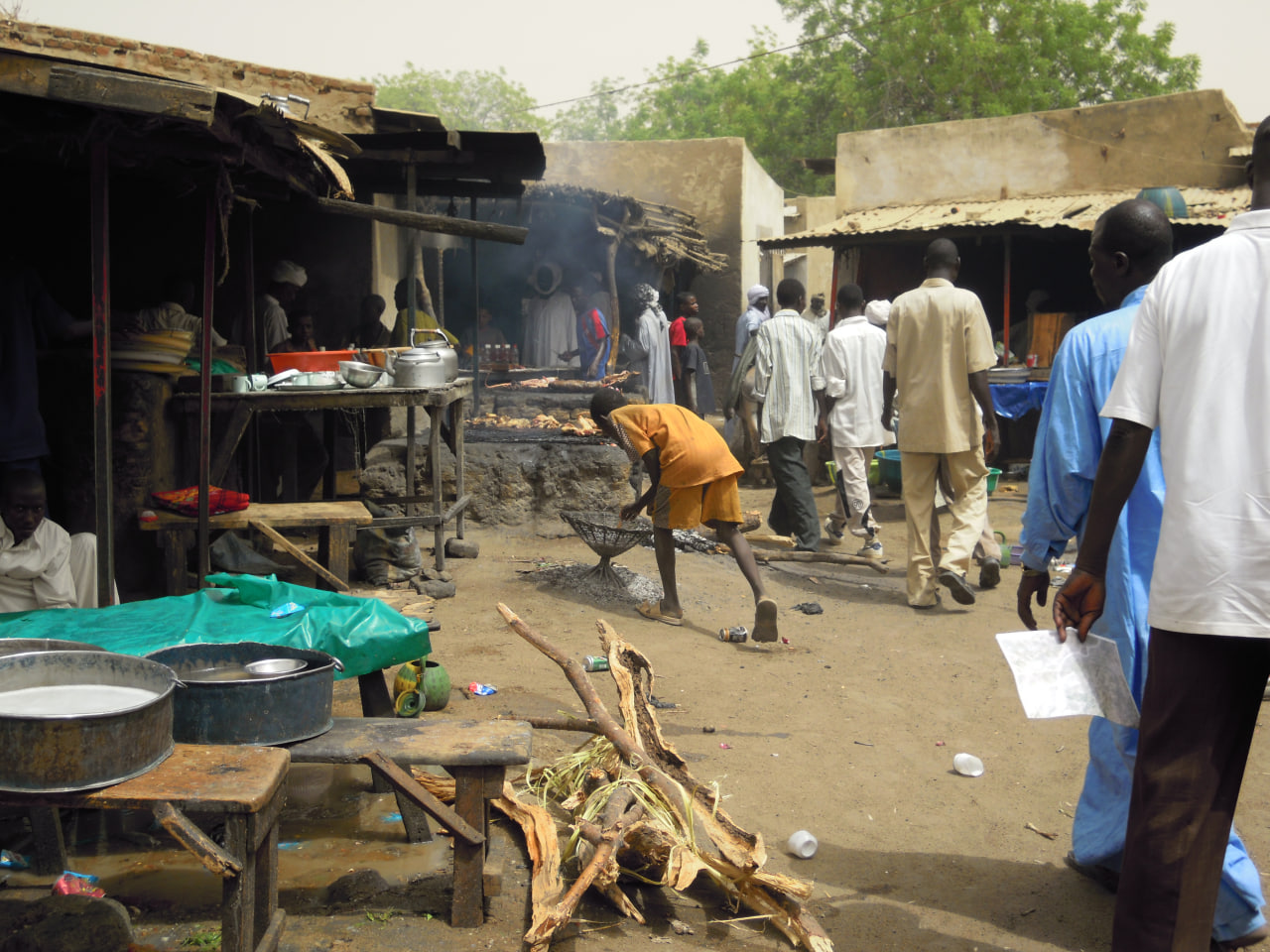}
    \includegraphics[width=.19\textwidth]{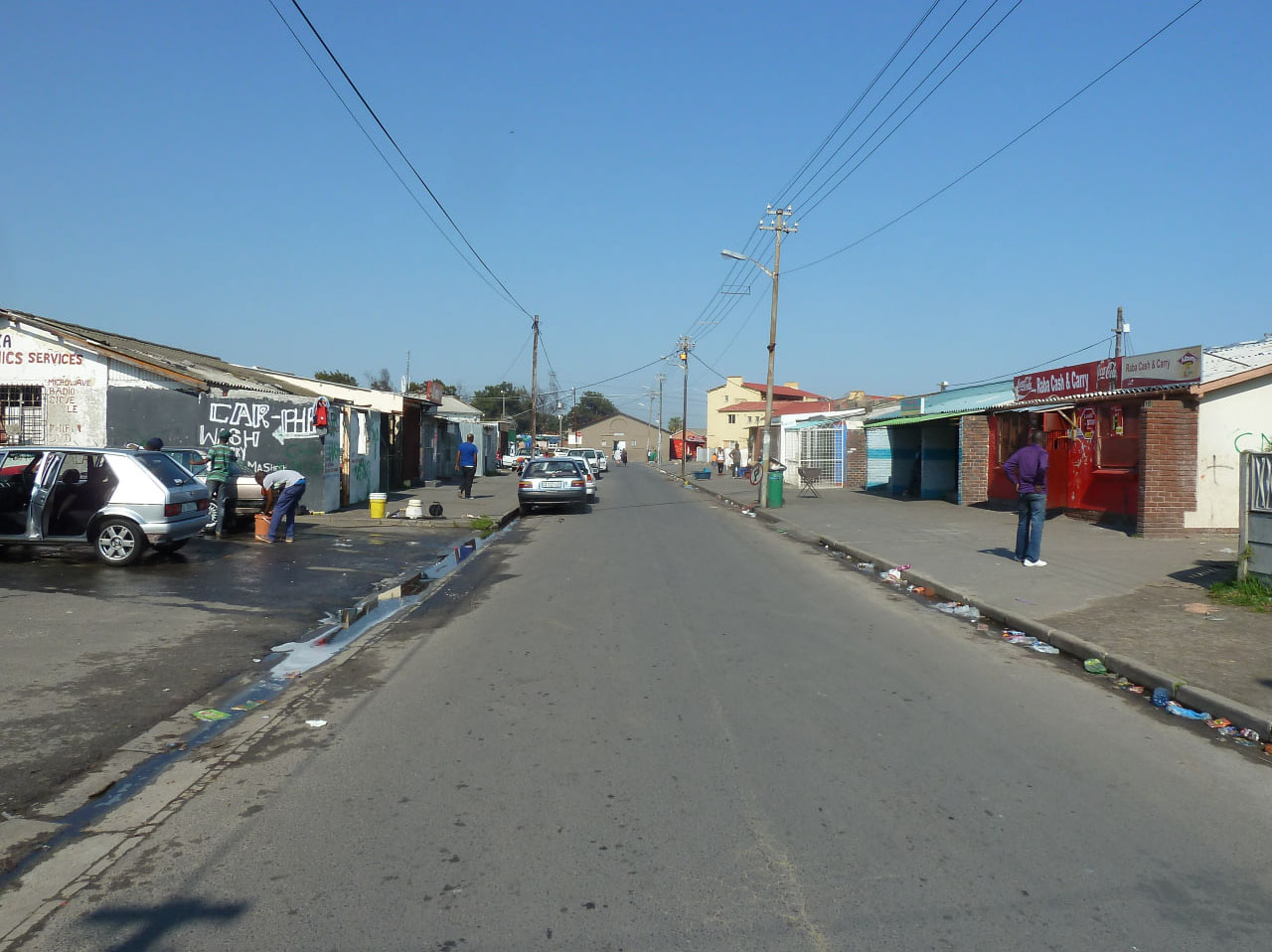}
    \includegraphics[width=.19\textwidth]{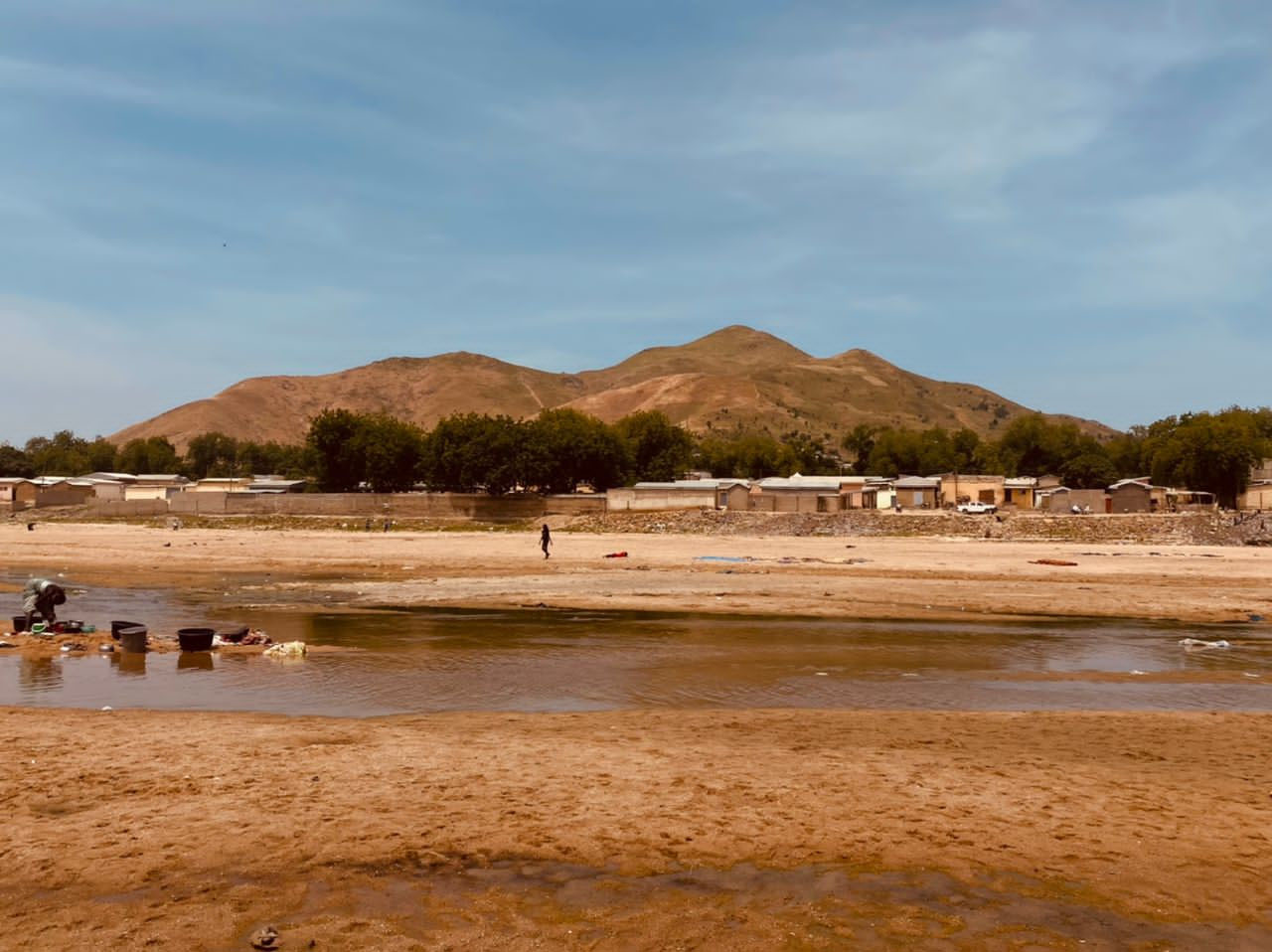}    
    \caption{\label{fig:sca100} SCA100}
  \end{subfigure}
  \caption{\label{fig:example-images} Example images from the IM2GPS \subref{fig:im2gps} and SCA100 \subref{fig:sca100} datasets.}
\end{figure}

We considered two datasets of geolocated images (\autoref{fig:example-images}):

\begin{itemize}
\item \textbf{IM2GPS3k} is a public benchmark dataset for image geolocation estimation evaluation \cite{vo2017revisiting} which features 3000 geolocated images from the broader IM2GPS dataset \cite{hays2008im2gps} (which consists of 6 million geo-tagged images collected from Flickr).
\item \textbf{SCA100} is a collection of 100 geolocated images from SenseCity Africa\footnote{Refer to: https://sensecity-africa.io (Visited on Jan. 22 2024)}, which is a crowd-sourced dataset aimed to explore the potential offered by communication technologies and the ubiquity of Internet access to address urban challenges in Africa. To ensure comprehensive representation across the continent, we employed proportional sampling factoring in the five major regions of Africa\footnote{Refer to: https://unstats.un.org/unsd/methodology/m49/\#geo-regions (Visited on Sep. 15 2023)}, namely North Africa, West Africa, East Africa, Central Africa, and Southern Africa. Then, the designated number of photos was randomly sampled from each region.
\end{itemize}

\begin{figure}[H]

  \centering
  \begin{subfigure}[b]{0.5\textwidth}
    \includegraphics[width=\textwidth]{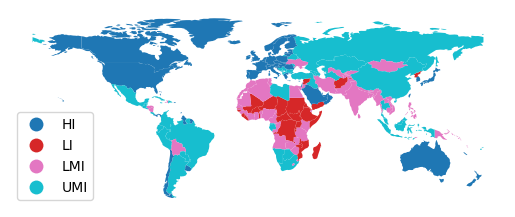}
    \caption{\label{fig:income-groups} Income groups}
  \end{subfigure}%
 \begin{subfigure}[b]{0.5\textwidth}
    \includegraphics[width=\textwidth]{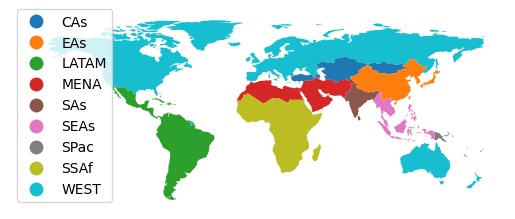}
    \caption{\label{fig:world-regions} World regions}
  \end{subfigure}

  \caption{\label{fig:regions} Contextual geographic divisions of the world. The map on the left (\subref{fig:income-groups}) illustrates the delineation of income groups according to the World Bank Group country classifications.  The one on the right (\subref{fig:world-regions}) shows the division of the world into nine distinct regions based on Jones' classification system.}
\end{figure}

In order to further investigate the geographic patterns of the prediction errors of the ISNs model, we included the following contextual data (\autoref{fig:world-regions}):

\begin{itemize}
    \item \textbf{Income groups}: following the World Bank Group country classifications \cite{world2022world}, i.e., low income (LI), lower middle income (LMI), upper middle income (UMI) and high income (HI).
    \item \textbf{World regions}: based on Jones \cite{jones2022unique}, which combines demographic, cultural, political, infrastructure, and geography aspects to classify countries into 9 unique regions of the world, i.e., the North and Australasia (WEST), Latin America (LATAM), Middle East and North Africa (MENA), Sub-Saharan Africa (SSAf), Central Asia (CAs), South Asia (SAs), East Asia (EAs), Southeast Asia (SEAs) and South Pacific (SPac).
\end{itemize}

\section*{Experiments and results}

\subsection*{Image geolocation estimation accuracy at different scales}


The accuracy of the ISNs model at each scale for the IM2GPS3k and SCA100 datasets is shown in \autoref{tab:isns_accuracy}.

\begin{table}[th]
     \centering
    \begin{tabular}{ p{1.7cm}p{1.7cm}p{1.7cm}  }

  \toprule
  \textbf{ Distance} &  \textbf{IM2GPS3k} & \textbf{SCA100} \\
  \midrule
  1 km   &  9.7 &   1.0\\
  25 km &  27.0   & 6.0\\
  200 km & 35.6 &  12.0\\
  750 km   & 49.2 &  18.0\\
  2500 km & 66.0 & 37.0\\

  \bottomrule
 \end{tabular}
     \caption{ISNs accuracy on IM2GPS3k and SCA100}
     \label{tab:isns_accuracy}
   \end{table}

   When comparing to the IM2GPS3k and SCA100, the ISNs model systematically obtained significantly lower accuracies for SCA100 at all the scales.
   At the street level (1 km), the ISNs model predicted 9.7\% of the IM2GPS3k images correctly, whereas only 1\% accuracy was obtained for SCA100.
   The accuracy for the IM2GPS3k dataset progressively increased to 27.0, 35.6, 49.2 and 66\% at the city (25 km), region (200 km), country (750 km) and continent (2500 km) levels respectively.
   In contrast, only 6\% of the SCA100 images were correctly predicted at the city level, and the accuracy respectively increased to 12, 18 and 37\% at the region, country and continent levels.


\subsection*{Confusions accross income groups and world regions}

\begin{figure}[H]

  \centering
  \begin{subfigure}{.49\linewidth}
    \centering
    IM2GPS3k
  \end{subfigure}
  \hfill
  \begin{subfigure}{.49\linewidth}
    \centering
    SCA100
  \end{subfigure}
  \begin{subfigure}[b]{\linewidth}
    \includegraphics[width=.49\linewidth]{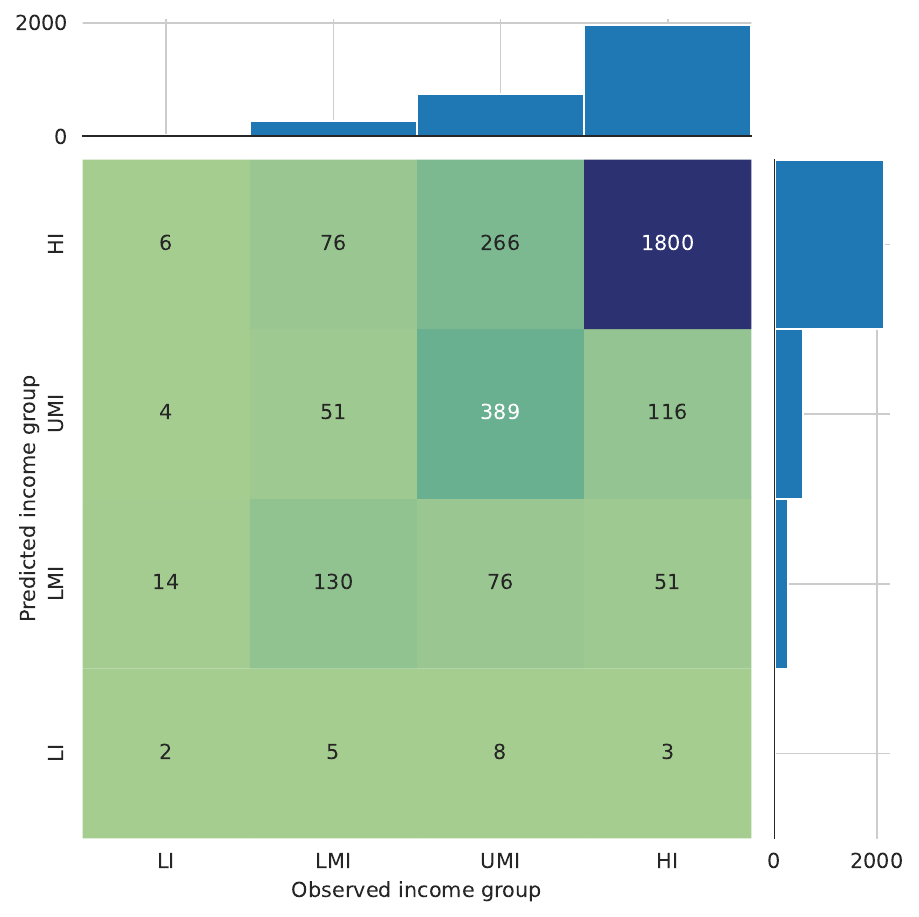}
    \includegraphics[width=0.49\linewidth]{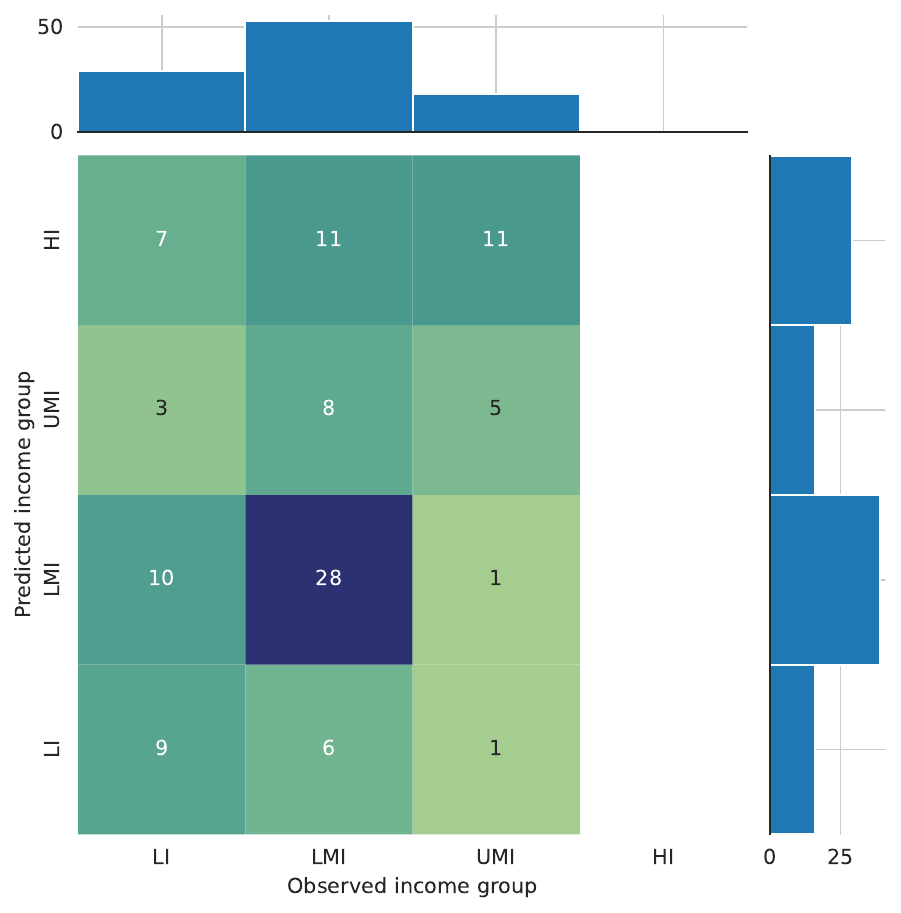}
    \caption{\label{fig:confusion-income} Income groups}
  \end{subfigure}
  \par\bigskip
  \begin{subfigure}[b]{\linewidth}
    \includegraphics[width=.49\linewidth]{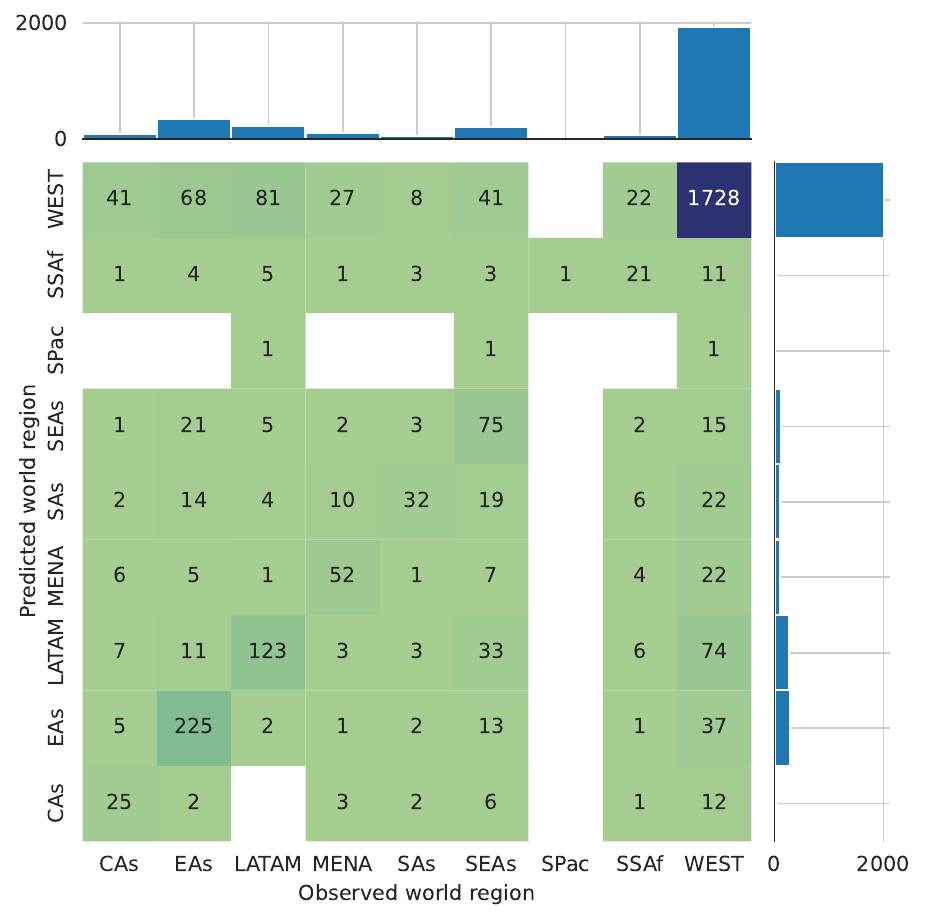}
    \includegraphics[width=.49\linewidth]{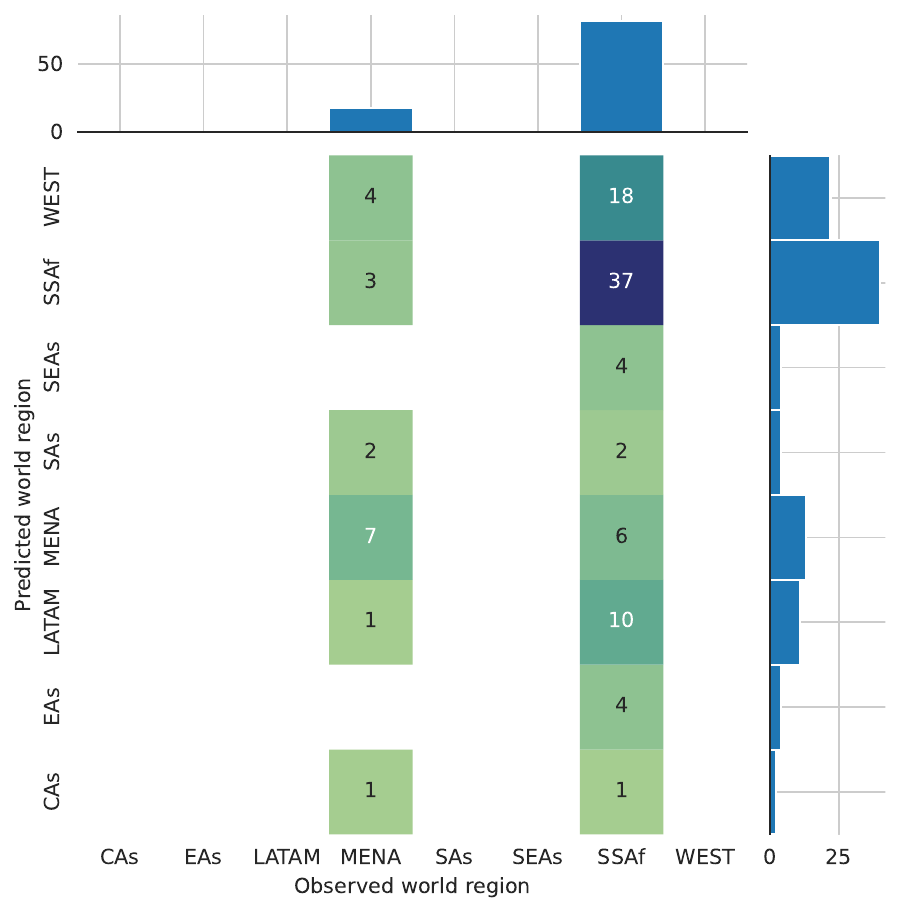}
    \caption{\label{fig:confusion-region} World regions}
  \end{subfigure}
  \caption{\label{fig:overall-confusion}Confusion matrices by world regions (\subref{fig:confusion-region} and income groups (\subref{fig:confusion-income}) for the IM2GPS3k (left) and SCA100 (right) datasets. The bars over each axis show the row (predicted) and column (observed) totals.}
\end{figure}

In order to explore the causes of the lower accuracy observed for the SCA100 dataset, we classified the images of the IM2GPS3k and SCA100 datasets into income groups and world regions, both for the original image locations and the locations predicted by ISNs.
We then used this classification to compute confusion matrices for the income groups and world regions (\autoref{fig:confusion-income} and \autoref{fig:confusion-region} respectively).

Regarding the distribution the images across income groups based on their original location, the majority of the IM2GPS3K images are situated in HI and UMI regions, accounting for 65.73 and 24.66\%  respectively, whereas only a 8.70 and 0.87\% represents LMI and LI regions respectively.
In contrast, no images of HI regions are originally found in SCA100, which consists of a 18, 53 and 29\% of images of  UMI, LMI, and LI regions respectively.

For IM2GPS3k, 91.37\% of the images originally in HI regions are correctly predicted in HI regions as well --- a figure which decreases to 52.64, 49.61 and 7.69\% for images correctly predicted in UMI, LMI and LI respectively. In fact, a 35.99, 29.01 and 23.08\% of images originally found in UMI, LMI and LI regions respectively are predicted in HI regions. The former are the most common confusions between income groups in relative terms after the a 53.85\% of images from LI regions that are predicted in LMI regions. The confusions have a clear tendency towards predicting images in higher income groups, with 92.23, 48.47 and 35.99\% of the images originally in LI, LMI and UMI regions respectively predicted in regions of higher income. In sharp contrast, only a 8.63, 11.37 and 1.91\% of images with actual locations respectively in HI, UMI and LMI regions are predicted in lower income regions.

The tendency of the ISNs model to predict image locations in income regions higher than the original location is also observed in the SCA100 dataset.
Although SCA100 does not feature any image located in HI regions, 29\% of the SCA100 images are predicted in HI regions.
The proportion of images originally in UMI, LMI and LI regions that are correctly predicted within its income group are of 27.78, 52.83 and 31.03\% respectively. The most common confusions between income groups in relative terms are a 61.11\% of images originally from UMI regions but predicted as HI regions, followed by a 34.48 and a 24.14\% of images originally in LI regions but respectively predicted as LMI and HI.
Finally, 68.97, 35.85 and 61.11\% of the images from LI, LMI and UMI regions respectively are predicted in regions of higher income, contrasting with the 11.11 and 11.32\% of images respectively located in UMI and LMI regions whose location is predicted in regions of lower income.

When considering the world regions of \autoref{fig:world-regions}, the IM2GPS3k also exhibited a strong bias towards countries in the Global North.
As depicted in \autoref{fig:confusion-region}, IM2GPS3k's images were predominantly located in the WEST region (64.13\%), with a notably smaller presence in East Asia (EAs) (11.67\%), LATAM (7.41\%), MENA (3.30\%), and SSAf (2.10\%).
On the other hand, by its own definition, SCA100 exclusively comprises pictures from SSAf (82\%) and MENA (18\%).
The ISNs predictions by world regions displayed an akin pattern to that of income groups, with a strong bias towards the WEST region. For example, in the IM2GPS3K dataset, a 46.59\% of images from Central Asia (CAs) were incorrecly predicted as WEST. The same was observed for 36.48, 34.92 and 27.27\% of the images from LATAM, SSAf and MENA respectively.
Similarly, in the SCA100 dataset, while a 45.12\% of the SSAf images and 38.88\% of the MENA were predicted correctly, a 21.95 and 22.22\% of the images from these regions were respectively predicted as WEST.

\subsection*{Clustering images by geolocation prediction error}

\begin{figure}[H]
  \centering
   \begin{subfigure}[b]{.49\linewidth}
    \includegraphics[width=\linewidth]{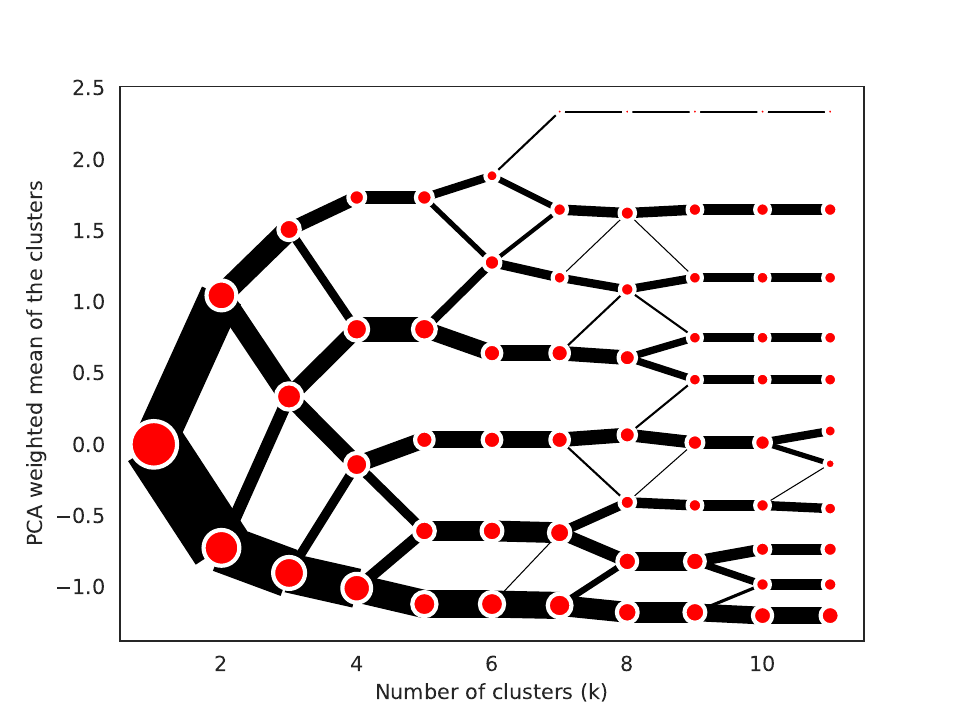}
    \vspace{-1em}
    \caption{\label{fig:clustegram}}
    \end{subfigure}
    \hfill
    \begin{subfigure}[b]{0.49\linewidth}
    \includegraphics[width=\linewidth]{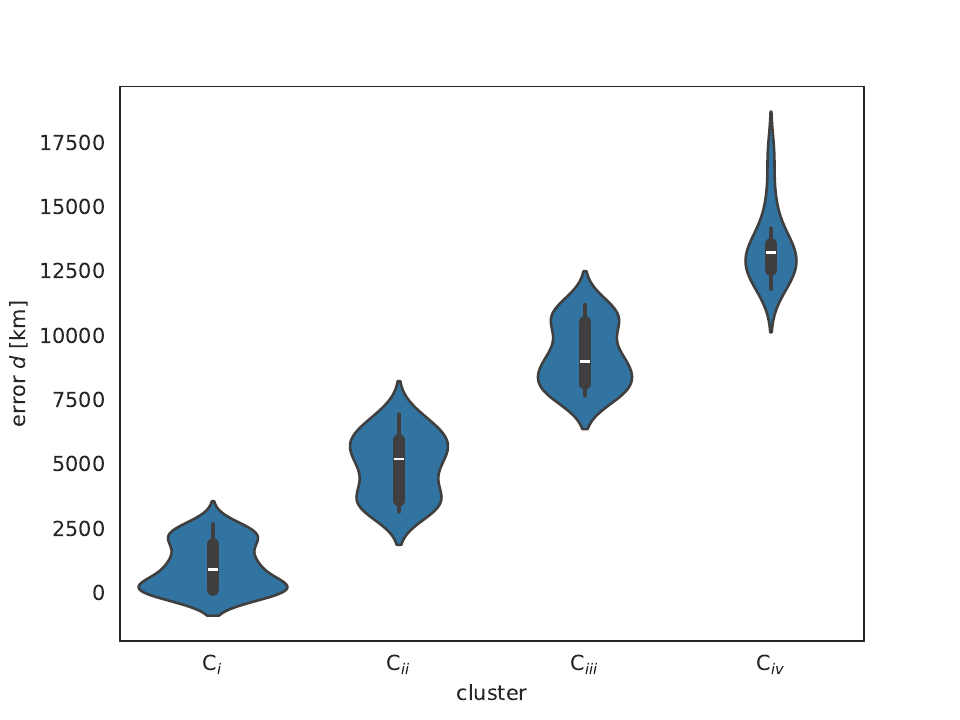}
    \vspace{-1em}
    \caption{\label{fig:distance-cluster-violin}}
  \end{subfigure}
  \vspace{-1em}
  \caption{\label{fig:clusters} Clustegram diagram (\subref{fig:clustegram}) based on the distance $d$ between the predicted and the ground truth location in the SCA100 dataset. Diagram obtained using the Python library clustergram \cite{fleischmann2023clustergram}. Violin plot (\subref{fig:distance-cluster-violin}) of the obtained clusters.}
\end{figure}

In order to explore the significantly lower accuracy in the SCA100 dataset, we first cluster its images according to the distance between the observed $y$ and predicted $\hat{y}$ photo locations, i.e., the residuals $d = gcd(y, \hat{y})$ where $gcd$ is the great-circle distance. Clusters are obtained using the KMeans algorithm \cite{macqueen1967some,lloyd1982least} using the Python library scikit-learn \cite{pedregosa2011scikit}.
Based on visual interpretation of the clustergram diagram proposed by Schonlau \cite{schonlau2002clustergram} (\autoref{fig:clustegram}), the number of clusters was set to $k=4$. 


The following clusters were obtained (\autoref{fig:distance-cluster-violin}): C$_i$ with 38 photos where $d < 2664$~km, C$_{ii}$ with 25 photos where $d \in [3145, 6929]~km$, C$_{iii}$ with 24 photos where $d \in [7655, 11185]$~km and C$_{iv}$ with 13 photos where $d > 17000.93$~km.
Notably, the first cluster already exceeds the continental scale (2500 m) threshold --- the broadest level used to evaluate image geolocation estimation accuracy (\autoref{tab:isns_accuracy}).

\begin{figure}[H]

  \centering
  \begin{subfigure}[c]{.85\textwidth}
    \raisebox{6em}{C$_{i}$}
    \hspace{.05\textwidth}%
    \includegraphics[width=.36\textwidth]{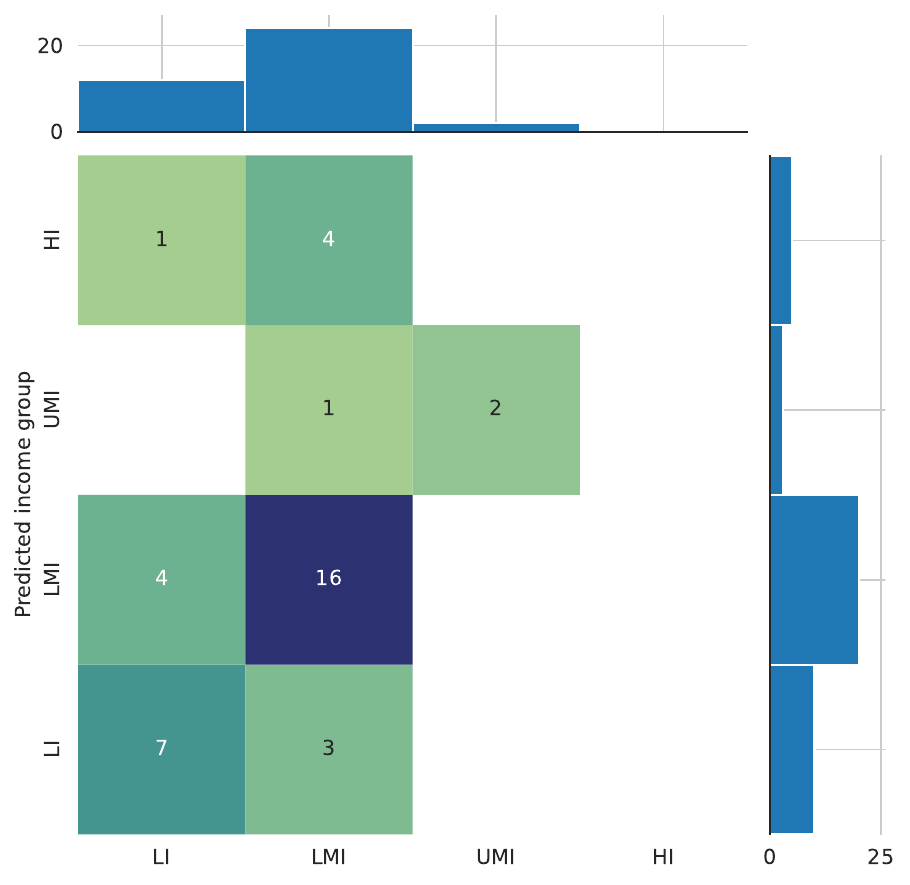}
    \hspace{.1\textwidth}%
    \includegraphics[width=.36\textwidth]{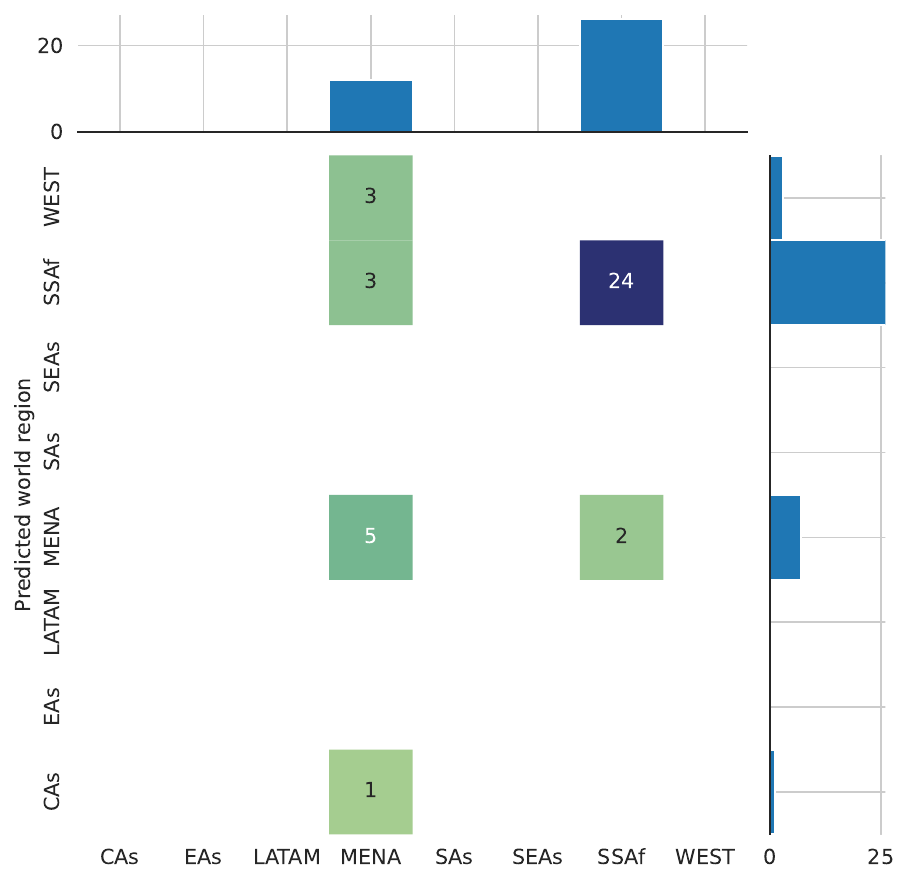}
  \end{subfigure}
  \begin{subfigure}[c]{.85\textwidth}
    \raisebox{6em}{C$_{ii}$}
    \hspace{.05\textwidth}%
    \includegraphics[width=.36\textwidth]{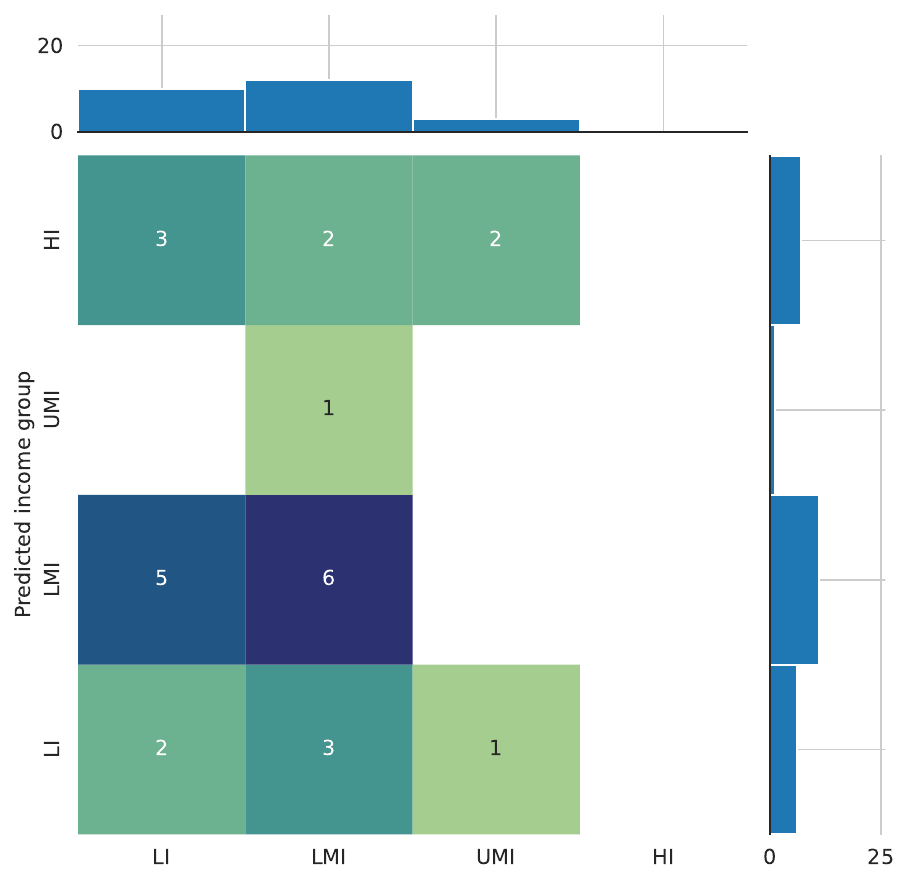}
    \hspace{.1\textwidth}%
    \includegraphics[width=.36\textwidth]{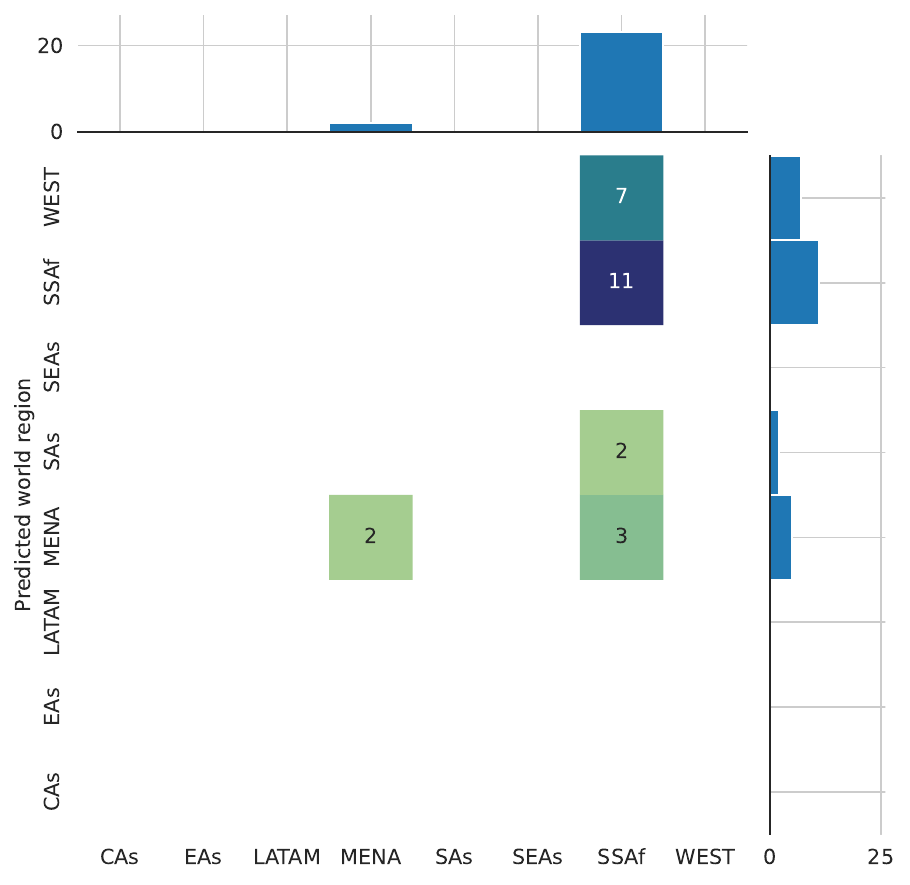}
  \end{subfigure}
  \begin{subfigure}[c]{.85\textwidth}
    \raisebox{6em}{C$_{iii}$}
    \hspace{.05\textwidth}%
    \includegraphics[width=.36\textwidth]{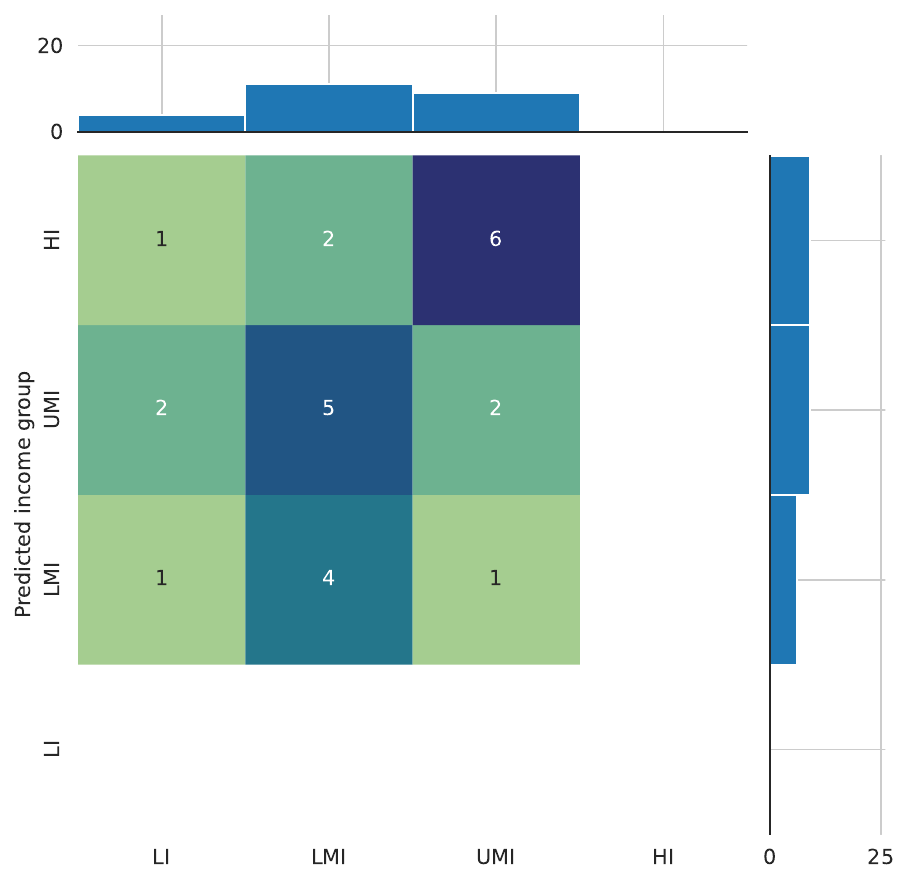}
    \hspace{.1\textwidth}%
    \includegraphics[width=.36\textwidth]{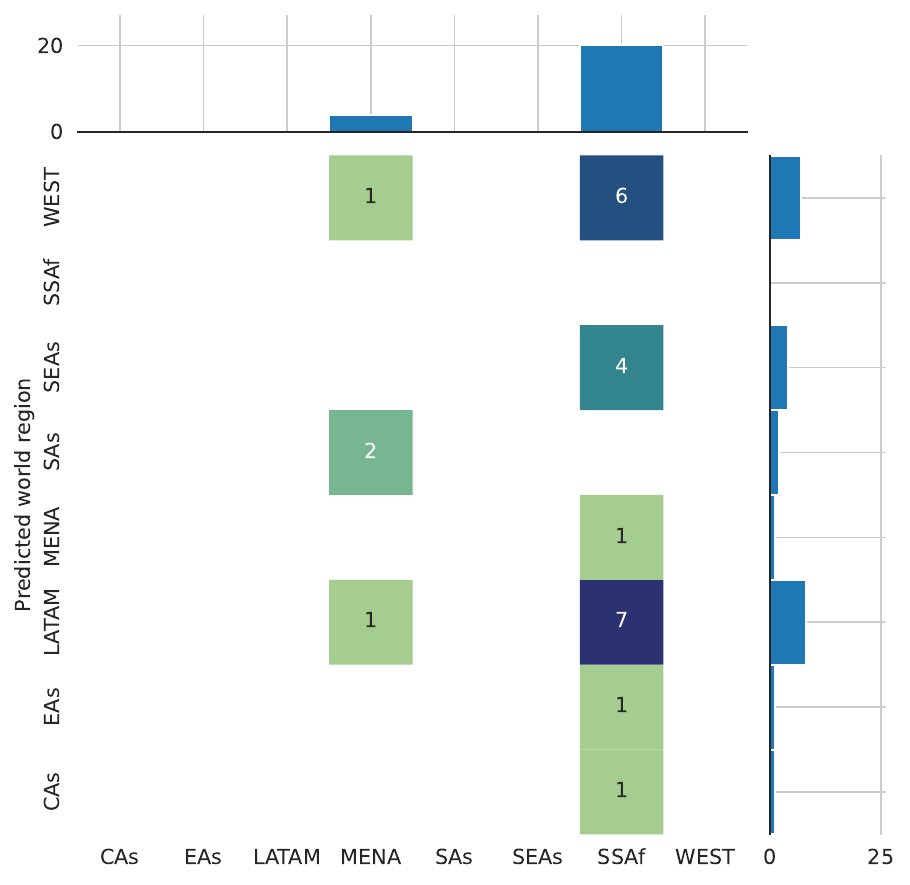}
  \end{subfigure}
  \begin{subfigure}[c]{.85\textwidth}
    \raisebox{6em}{C$_{iv}$}
    \hspace{.05\textwidth}%
    \includegraphics[width=.36\textwidth]{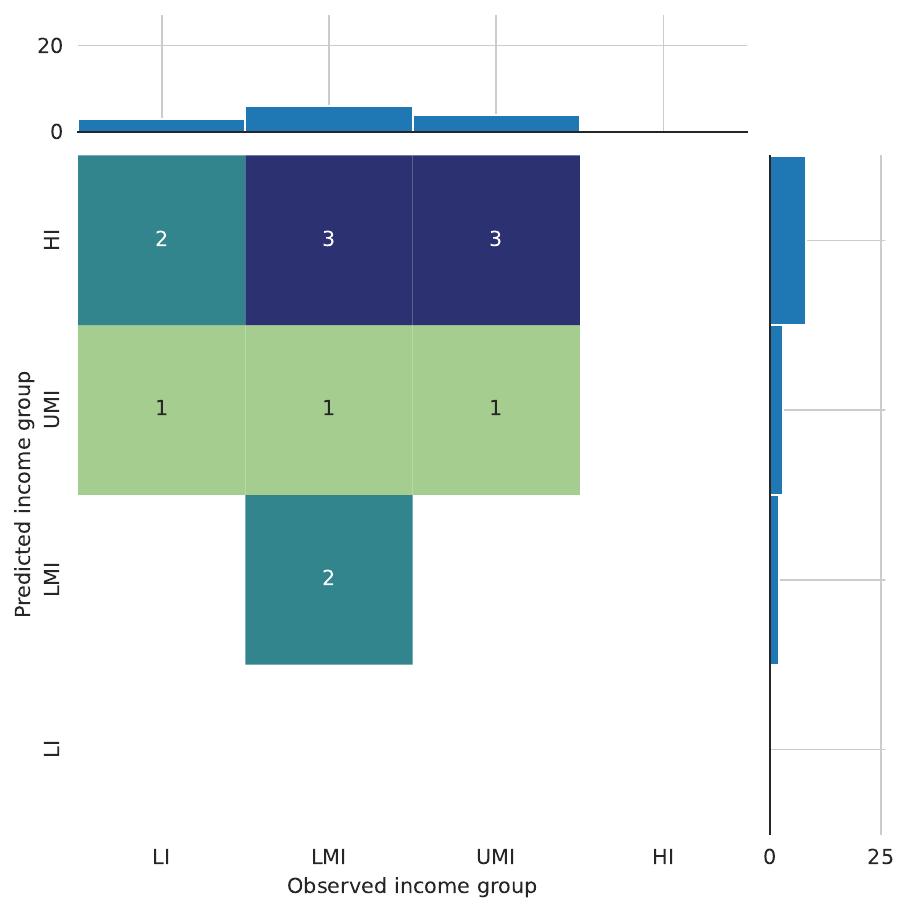}
    \hspace{.1\textwidth}%
    \includegraphics[width=.36\textwidth]{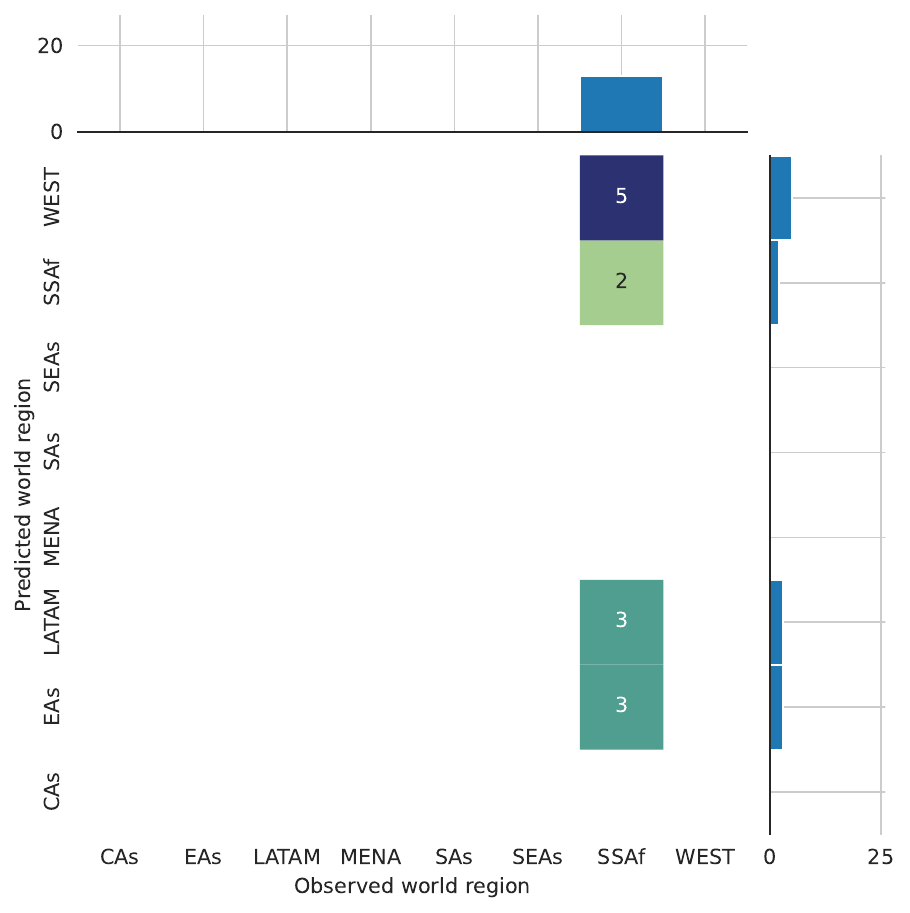}
  \end{subfigure}
  \begin{minipage}{.85\textwidth}
    \phantom{C$_{iii}$}%
    \hspace{.05\textwidth}%
    \begin{subfigure}[c]{.36\textwidth}
      \caption{\label{fig:cluster-confusion-income} Income group}
    \end{subfigure}%
    \hspace{.1\textwidth}
    \begin{subfigure}[c]{.36\textwidth}
      \caption{\label{fig:cluster-confusion-region} World region}
    \end{subfigure}
  \end{minipage}

  \caption{\label{fig:cluster-confusions}Confusion matrices for the SCA100 clusters C$_i$ to C$_{iv}$ (top to bottom), for income groups (\subref{fig:cluster-confusion-income}) and world regions (\subref{fig:cluster-confusion-region}).}
\end{figure}

In order to further explore the lower accuracy observed for the SCA100 dataset, \autoref{fig:cluster-confusions} presents the confusion matrices of the ISNs model for each cluster C$_i$ to C$_{iv}$, income group and world region.

The pattern in which the ISNs tends to predict image locations in higher income groups is exacerbated in clusters with higher prediction error $d$ --- this can be visually noted since the highest values of the confusion matrices are increasingly found in the upper-left parts as we move from C$_i$ to C$_{iv}$ (from top to bottom in \autoref{fig:cluster-confusion-income}).
The proportion of images predicted in their correct income group decreases from 65.79, 32.00, 25.00 and 23.08\% in C$_i$, C$_{ii}$, C$_{iii}$ and C$_{iv}$ respectively, whereas the proportion of images predicted in an income region higher than its original location respectively increases from 26.32, 52.00, 70.83 and 76.92\% in each cluster.
Additionally, it can be observed that in the clusters with higher prediction error $d$ --- C$_{iii}$ and C$_{iv}$ --- none of the images originally located in LI regions are predicted as such.

When considering the confusion matrices for world regions accross clusters (\autoref{fig:cluster-confusion-region}), an analogous pattern can be noted in which the bias towards mistakenly predicting image locations in the WEST region increases from clusters C$_i$ to C$_{iv}$ (hence increasing prediction error $d$).
In C$_i$, most of the photos originally located at SSAf are correctly predicted within the region (92.30\%) and the others are predicted as MENA (7.70\%). At the same time, 41.67\% of the MENA images are predicted as such, whereas a 25\% are predicted as WEST and another 25\% as SSAf, with the remaining 8.33\% are predicted as CAs.
The second cluster C$_{ii}$ is comprised almost exclusively of SSAf images, a 47.82\% of which are correctly predicted within the region, yet a 30.43\% are predicted as WEST and a 13.04 and 8.69\% are predicted as MENA and SAs respectively. The only images from the MENA region are correctly predicted.
The third cluster C$_{iii}$ shows a surprising pattern as it is mostly composed of images from SSAf yet not a single one is predicted as such. A 14, 12 and 8\% of the photos are predicted as LATAM, WEST and SEAs respectively, while the remaining three photos are predicted as CAs, EAs and MENA.
Similarly, none of the images originally located in MENA is correctly predicted within the region, with two predicted as SAs, one predicted as LATAM and another as WEST.
Finally, C$_{iv}$, the cluster with highest $d$, consists solely of images from SSAf, among which a 38.46, 23.07 and 23.07\% are predicted in WEST, EAs and LATAM regions respectively and only 15.38\% are correctly predicted as SSAf.

\section*{Discussion and conclusion}
Our study employed the ISNs image geolocation estimation model to assess its accuracy using two distinct datasets --- IM2GPS, a global dataset but consistingl mostly of images from the Western world and SCA100, a dataset comprising only photos from the African continent (a strongly underrepresented region in the first dataset).
Our primary objective was to identify inherent regional biases within the ISNs model and its training dataset, namely the IM2GPS.
Through a careful comparison of the outcomes from IM2GPS3K and SCA100, our findings reveal a pronounced bias favoring regions in the global north.
Further test delved into the nature of these biases and revealed a significant inclination towards high income countries.
As a result, the ISNs model, performed remarkably poorly when confronted with data from SCA100.

Ensuring a balanced presence of social and cultural nuances in the training data is crucial so that citizens from underrepresented regions can also benefit from continuous advances in AI and CV. Image geolocation estimation holds strong potential for numerous applications such as navigation, disaster management or assisting archiving collections of historical images in antrhopological studies in which the image location is a central attribute \cite{scherer1990historical,banks2010introduction}, and the availability of data in developing regions such as Africa is scarce \cite{geary1986photographs,schneider2010topography,chenal2013ville}.

The results presented in this study show a strong relationship between the intrinsic geographic characteristics of the training data and the model outputs, which challenges the appropriateness of the current datasets and benchmarks in image geolocation estimation. However, it is important to acknowledge the limitations of our work. First, it considered only biases related to income groups and conceptual world regions, and it was restricted to a single model and dataset. Another notable shortcoming is the small size of our SCA100 dataset which in turn, highlights the difficulty of obtaining high quality training data in underrepresented regions - the initial motivation of this study.
Therefore, future work will focus on collecting larger datasets of underrepresented regions, applying the approach to evaluate regional biases to further image geolocation estimation datasets and models and ultimately use the gained insights to fine-tune existing models so that they can better serve the needs of citizens in underrepresented regions.

\section*{Acknowledgments}
This research has been supported by the École Polytechnique Fédérale de Lausanne (EPFL).


\bibliography{bibliography}

\bibliographystyle{abbrv}

\end{document}